\documentclass[sigconf]{acmart}

\usepackage{makecell}
\usepackage{amsmath,amssymb,amsfonts}
\usepackage{amsthm}
\usepackage{multirow}
\usepackage{caption}
\usepackage{subcaption}
\usepackage{color} 
\usepackage{colortbl}
\usepackage{multirow}

\usepackage{lipsum}
\usepackage{enumitem}
\usepackage{graphicx}
\usepackage{lipsum}

\usepackage[framemethod=tikz]{mdframed}
\definecolor{mygray}{RGB}{243,243,244}
\newmdenv[
  innertopmargin=0pt,
  backgroundcolor=mygray,
  linecolor=none,
  innerleftmargin=0pt,
  innerrightmargin=0pt,
  leftmargin=0pt
  ]{mymath}

\usepackage[ruled, linesnumbered,vlined]{algorithm2e}

\usepackage{amsthm}
\newtheorem{theorem}{Theorem}
\newtheorem{definition}{Definition}

\AtBeginDocument{%
  }

\copyrightyear{2026}
\acmYear{2026}
\setcopyright{cc}
\setcctype{by}
\acmConference[WWW '26] {Proceedings of the ACM Web Conference 2026}{April 13--17, 2026}{Dubai, United Arab Emirates.}
\acmBooktitle{Proceedings of the ACM Web Conference 2026 (WWW '26), April 13--17, 2026, Dubai, United Arab Emirates}
\acmISBN{979-8-4007-2307-0/2026/04}
\acmDOI{10.1145/3774904.3792120}
\begin{document}

\title{Riemannian Graph Tokenizer for Structural Knowledge Transfer}

\author{Qimin Zhou}
\affiliation{%
  \institution{Beijing University of Posts and Telecommunications}
  \city{Beijing}
  \country{China}
}
\email{zhouqimin@bupt.edu.cn}

\author{Haibo Liu}
\affiliation{%
  \institution{Beijing University of Posts and Telecommunications}
  \city{Beijing}
  \country{China}
}
\email{lhb0410@bupt.edu.cn}

\author{Yujie Wang}
\affiliation{%
  \institution{Beijing University of Posts and Telecommunications}
  \city{Beijing}
  \country{China}
}
\email{yujiewang@bupt.edu.cn}

\author{Li Sun}
\authornote{Corresponding author.}
\affiliation{%
  \institution{Beijing University of Posts and Telecommunications}
  \city{Beijing}
  \country{China}
}
\email{lsun@bupt.edu.cn}

\author{Chuan Shi}
\authornotemark[1]
\affiliation{%
  \institution{Beijing University of Posts and Telecommunications}
  \city{Beijing}
  \country{China}
}
\email{shichuan@bupt.edu.cn}

\renewcommand{\shortauthors}{Qimin Zhou, Haibo Liu, Yujie Wang, Li Sun, \& Chuan Shi.}

\begin{abstract}
Foundation models are at the forefront of artificial intelligence. A tokenizer, converting the raw input into discrete representations that the model can understand, plays an important role to the success of foundation models. Unlike the text tokenizer that is well studied in large language models, graph tokenizer is still at its early stage, facing the challenges of tackling the non-Euclidean structures and capturing the structural semantics. \textbf{How to design a graph tokenizer for structural knowledge transfer?} To this end, we propose a Riemannian Graph Tokenizer (\textbf{RGT}) that bridges the structural knowledge and quantized representations to support cross-domain structural knowledge transfer. The connection is established by  Riemannian geometry. Specifically, we first define the geometric vocabulary (trees, cycles and sequences), which captures fundamental structural patterns and reflects the intrinsic geometry of graph. Second, we construct a Riemannian quantizer with Riemannian Straight-Through Estimator to tokenise graph structures across multiple domains into discrete tokens. To ensure consistency and transferability across diverse geometric spaces, RGT further incorporates a geometry-aligned decoder that projects manifold-specific tokens into a unified tangent space. The theoretical analysis and geometric interpretations are provided to support the effectiveness of our proposed method. Extensive experiments across diverse datasets demonstrate that RGT significantly enhances structural knowledge transferability across graph domains.
\end{abstract}




\maketitle

\section{Introduction}\label{sec1}

Graph is a universal language for describing and modeling complex systems~\cite{graphsurveyyang,wu2020comprehensive}. 
Graphs data are ubiquitous in the real world, such as molecular graph~\cite{gaudelet2021utilizing}, social graph~\cite{wu2025graphfoundationmodelsrecommendation} and citation networks~\cite{kipf2016semi}. Graph machine learning~\cite{xia2021graph} aims to harness machine learning techniques~\cite{alpaydin2021machine} to uncover latent information in graphs and enhance the performance of downstream tasks. Among various graph machine learning approaches, Graph Neural Networks (GNNs) ~\cite{hamilton2017inductive,veličković2018graphattentionnetworks,yan2024federated} have received increasing attention in recent years due to their broad applicability to various Web-related problems~\cite{ma2024hetgpt} and strong empirical performance~\cite{liu2025widegate, lu2021social}. However, most existing GNNs methods are typically tailored to specific datasets or tasks, lacking generalization across domains and tasks. Consequently, Graph Foundation Models (GFMs) ~\cite{DBLP:conf/icml/MaoCT000S0T24,he2025position,zhang2025can} are emerging as an interesting research topic in the graph domain.

GFMs~\cite{liu2025graph} refer to pretrained models across diverse graph datasets that support multiple downstream tasks. The GFMs proposed by researchers primarily address the aforementioned limitations of GNNs. Some methods leverage the message-passing mechanism of GNNs to aggregate information. All in One~\cite{DBLP:conf/ijcai/Sun000G24} proposes a novel multi-task prompting approach to unify downstream tasks. GCOPE~\cite{DBLP:conf/kdd/ZhaoCS0L24} further conducts model training on multi-domain graphs with coordinators in order to improve the generalization capacity to different  datasets. Recently, GraphAny~\cite{zhaofully} models inference on a new graph as an analytical solution to a LinearGNN, which can be naturally applied to graphs with any feature and label. Moreover, some methods align graph structural knowledge with textual spaces. A representative example is OFA~\cite{DBLP:conf/iclr/0057FKLT0Z24}, which encodes textual attributes using pretrained language models to support various graph tasks. ZeroG~\cite{DBLP:conf/kdd/0001WLY024} further enables zero-shot transfer across graph datasets by aligning semantic and structural features through textual descriptions of nodes and classes. In addition, GFT~\cite{DBLP:conf/nips/WangZCZ024} aligns the structural information of computation tree vocabulary with textual representations through a quantization-based approach. Notably, recent research has begun to explore the transfer of structural knowledge in graphs. For example, OpenGraph~\cite{DBLP:conf/emnlp/XiaK024} presents a unified graph tokenizer and scalable graph transformer to encode graph structures. It also utilizes large language models to enhance graph features.  RiemannGFM~\cite{DBLP:conf/www/0008HZW0Y25} learns structural knowledge solely through structural vocabulary, enabling cross-domain generalization without relying on textual features. GFMs have made remarkable progress, yet transferring structural knowledge across graphs remains a significant challenge.




In contrast to GFMs, the foundation models pretrained on large-scale text or image data~\cite{openai2024gpt4technicalreport,wang2023visionllm} have demonstrated properties such as emergence and homogenization. Their tokenizers~\cite{bostrom2020byte,kudo2018sentencepiece} serve as critical bridges between raw data and learned representations, significantly enhancing model generalization and cross-domain transferability~\cite{chang2024survey}. 
Compared with text or image, graph has richer structural information and topological heterogeneity~\cite{guo2025graphmore,sun2022self,sun2023deepricci}. GFT and OpenGraph demonstrate that graph tokenizer plays a critical role in addressing the challenge of GFMs. However, research on graph tokenizer remains in its infancy, facing significant challenges in handling multi-domain non-Euclidean structures and capturing structural semantics for effective structural knowledge transfer. Specially, one problem lies in the substantial structural differences among graphs from different domains. Existing models often lack the ability to model and recognize transferable structural patterns. While computation trees~\cite{chuang2022tree} and structural vocabulary~\cite{DBLP:conf/www/0008HZW0Y25} provide partial solutions, relying solely on trees and cycles proves insufficient to represent the full diversity of graph structures. Another key limitation is that current GFMs cannot effectively perform structure-level knowledge transfer across graph domains. Tokenizer-based mechanisms have shown promise in compressing semantics into discrete and transferable tokens in shallow spaces~\cite{van2017neural,wang2025learning}. But existing efforts~\cite{DBLP:conf/nips/WangZCZ024} either depend heavily on textual alignment to facilitate tokenization and structural transfer, or distill GNNs into lightweight MLPs~\cite{DBLP:conf/iclr/0006TXLHQZ0ZL24}. Despite these advancements, the potential of tokenizer for enabling structure-level knowledge transfer remains largely underexplored. This naturally leads to a key question: \textit{How can we design a graph tokenizer that supports the structural knowledge transfer?}


 In this paper, inspired by the success of tokenizers in large language models and the expressive power of Riemannian geometry~\cite{lee2006riemannian,sun2025deeper} in modeling non-Euclidean graph structures~\cite{DBLP:conf/gd/Sarkar11,gu2018learning,sun2024spiking,sun2024lsenet}. We propose a novel \textbf{R}iemannian \textbf{G}raph \textbf{T}okenizer (RGT) that quantizes graph structural features into discrete tokens embedded in constant curvature spaces to support cross-domain structural knowledge transfer. Specifically, we sample geometric vocabulary as geometric structural patterns, and utilize a Riemannian encoder to learn the coordinates of trees, cycles and sequences in hyperbolic~\cite{sarkar2011low}, spherical~\cite{petersen2006riemannian} and Euclidean~\cite{topping2021understanding} spaces, respectively.
The coordinates of the geometric vocabulary are quantized using codebooks. We develop a Riemannian straight-through estimator to support gradient-based optimization in Riemannian spaces. The node representations are then parallel transported to the tangent space at the pole. Finally, the decoder reconstructs the original structural features, while reconstruction loss and contrastive loss are adopted to encourage the learning of structurally transferable representations. 
Our main contributions are summarized as follows:
\begin{itemize}

\item \textbf{Geometric Vocabulary.} The geometric vocabulary (trees, cycles and sequences) serves as fundamental structural patterns that facilitate structure-level knowledge transfer. 

\item \textbf{Riemannian Graph Tokenizer.} We propose RGT that quantizes graph structures across multiple domains within constant curvature spaces. Among its components, the Riemannian quantizer preserves the intrinsic geometry of non-Euclidean manifolds and is supported by theoretical analysis.

\item \textbf{Extensive Evaluation.} We conduct thorough theoretical analysis and experiments on multiple real-world graph datasets, showing that RGT achieves competitive performance on graph tasks and exhibits significant cross-domain generalization.
\end{itemize}

\section{Preliminaries}\label{sec2}

\paragraph{Riemannian Geometry.}\label{2.1}
For a smooth manifold $\mathcal{M}$, each point $\boldsymbol{x} \in \mathcal{M}$ is associated with a tangent space $T_{\boldsymbol{x}}\mathcal{M}$, which serves as the first-order approximation of the manifold near $\boldsymbol{x}$ and locally exhibits Euclidean structure. A Riemannian metric $\mathfrak g_{\boldsymbol{x}}(\cdot,\cdot): T_{\boldsymbol{x}}\mathcal{M} \times T_{\boldsymbol{x}}\mathcal{M} \to \mathbb{R}$ is defined on the tangent space $T_{\boldsymbol{x}}\mathcal{M}$, endowing it with an inner product. This allows geometric concepts such as lengths, angles, and distances to be meaningfully defined on the manifold. The tuple $(\mathcal{M}, \mathfrak{g})$ is thus called a Riemannian manifold. The curvature $\kappa_{\boldsymbol x}$ measures the deviation of the manifold at point $\boldsymbol x$ from flat Euclidean space. If a manifold has constant curvature $\kappa_{\boldsymbol x} = \kappa$ for all $\boldsymbol x \in \mathcal{M}$, it is called a Constant Curvature Space (CCS). 

\paragraph{Constant Curvature Space.} We unify the Lorentz, Spherical and Euclidean space under a unified $\kappa$-CCS space, denoted by 
\begin{equation}\label{eq1} \resizebox{0.9\linewidth}{!}{$\mathcal{L}_\kappa^d := \begin{cases} \left\{ \boldsymbol{x} = [ x_t;\boldsymbol x_s]\in \mathbb{R}^{d+1} \;\middle|\; \kappa \langle \boldsymbol{x}, \boldsymbol{x} \rangle_\kappa = 1,\; x_t > 0,\; \boldsymbol{x}_s \in \mathbb{R}^{d} \right\}, & \kappa \neq 0, \\ \left\{ \boldsymbol{x} = [x_t;\boldsymbol x_s] \in \mathbb{R}^{d+1} \;\middle|\; \langle \boldsymbol{x}, \boldsymbol{x} \rangle_\kappa \geq0, \boldsymbol x_t=0, \boldsymbol{x}_s \in \mathbb{R}^{d}\right\}, & \kappa = 0, \end{cases}$} \end{equation}
where the inner product is defined as:
$\langle \boldsymbol{x}, \boldsymbol{y} \rangle_\kappa := \operatorname{sgn}(\kappa) x_t y_t + \boldsymbol{x}_s^\top \boldsymbol{y}_s,$ $d$ and $\kappa$ denote the dimension and curvature. The Riemannian metric at $\boldsymbol x$ is $\mathfrak{g}_x= \operatorname{diag}(\operatorname{sgn}(k), 1, \ldots, 1)$. $x_t$ corresponds to the axis of symmetry of the hyperboloid and is termed the time-like dimension, while $\boldsymbol{x}_s \in \mathbb{R}^d$ denotes the space-like dimensions. This formulation recovers the following standard models under different curvature regimes. $\kappa < 0$: corresponds to the \textbf{Lorentz space} of hyperbolic geometry. $\kappa > 0$: corresponds to the \textbf{Spherical space}. $\kappa = 0$: corresponds to the \textbf{Euclidean space}. We define the “north pole” of the manifold as: $\boldsymbol o_\kappa =[\frac{1}{\sqrt{|\kappa|}}, 0, \cdots, 0]^\top.$ \textbf{The tangent space at the north pole} is given by $T_{\boldsymbol{o}_\kappa} \mathcal{L}_\kappa^d = \left\{ \boldsymbol{v} = \left[ 0; \boldsymbol{v}_s \right] \in \mathbb{R}^{d+1} \right\} \cong \mathbb{R}^d.$ Closed-form
exponential and logarithmic maps exist. 

\paragraph{Notations \& Problem Formulation.}\label{2.2}
A graph is defined as $ \mathcal{G} = (\mathcal{V}, \mathcal{E}, \mathcal{D}) $, where $ \mathcal{V}$ denotes the set of nodes, $ \mathcal{E} \subseteq \mathcal{V} \times \mathcal{V}$ denotes the set of edges and $\mathcal{D}$ denotes the domain. Each node $ v \in \mathcal{V}$ may optionally be associated with an attribute vector  $\boldsymbol{x}_v \in \mathbb{R}^d$. The graph tokenizer $ \mathcal{T}_\kappa$ is pretrained on multi-domain graph $ \mathcal{G} $ via self-supervised learning. It is parameterized by $ \boldsymbol{\theta}$ and associated with a discrete codebook $ \mathcal{C} = \{\boldsymbol c_1,\boldsymbol c_2, \dots,\boldsymbol c_m\}$ defined in a constant curvature space, where $\kappa$ denotes the curvature of the space. Formally, given a subgraph or structural pattern $ \mathcal{G}_s \subseteq \mathcal{G}$, the pretrained tokenizer encodes graph features for downstream tasks: $\mathcal{T}_\kappa: \mathcal{G}_s \mapsto \boldsymbol q_s \in \mathcal{C},$ where the tokenizer maps the input structural pattern to a geometry-aware token $\boldsymbol q_s $ selected from the codebook $\mathcal{C}$, which lies on the manifold $\mathcal{L}_\kappa^d $.

The proposed graph tokenizer is designed with two core capabilities. \textbf{Structure Awareness:} The model is expected to learn generalizable representations of fundamental structural patterns such as trees, cycles and sequences. \textbf{Structural
Knowledge Transfer:} The model should be pretrained on graph data from multiple domains and adapted with minimal parameter updates to new graph domains with varying structural or attribute distributions, thereby enabling generalized graph representations across domains.
\section{RGT: Riemannian Graph Tokenizer for Structural
Knowledge Transfer}\label{sec3}

\begin{figure*}[tbp]
  \vspace{-0.1in}
  \centering
\includegraphics[width=1\linewidth]{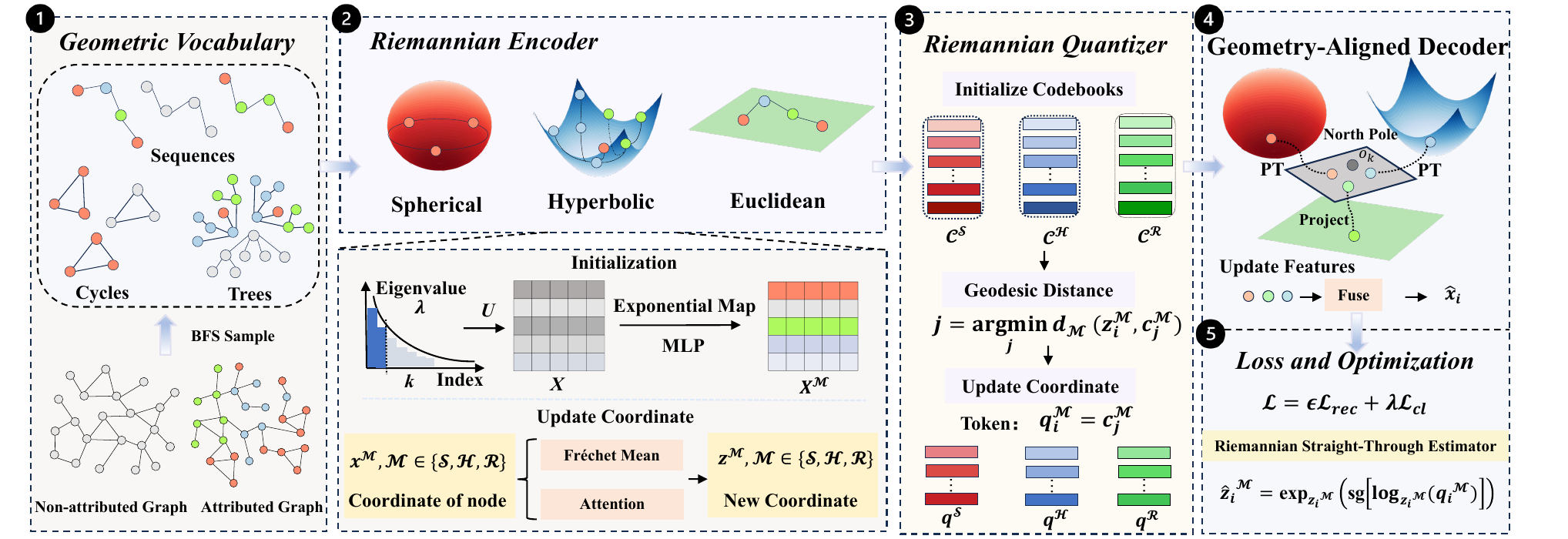}
  \caption{Overall architecture of the proposed Riemannian Graph Tokenizer.}
  \vspace{-10pt}
  \label{fig:architecture}
\end{figure*}

This section introduces the proposed RGT in detail, which is tailored for self-supervised pretraining on graph structures. The model discretizes structural features into geometry-aware representations. It supports both attributed and non-attributed graphs, enabling effective cross-domain transfer. Definition~\ref{def1} formalizes a complete geometric vocabulary that supports the patterns of arbitrary graph structures.
\begin{definition}[Geometric Vocabulary]\label{def1}
A set of graph substructures $S = \{S_1, S_2, \dots, S_K\} $ is called \emph{geometric vocabulary} if it satisfies the following two conditions:
\begin{itemize}
  \item \textbf{Structural Completeness:}
  For any graph \( \mathcal{G} = (\mathcal{V}, \mathcal{E}) \), there exists a multiset of substructures \( \{S_i^{(j)}\}_{j=1}^M \), where \( S_i^{(j)} \in S \), such that: $\mathcal{G} = \bigcup_{i=1}^K \bigcup_{j=1}^M S_i^{(j)}$. The value of $M$ can vary across different $S_i$, as each structural type may contribute a different number of instances in the composition of the graph.
  \item \textbf{Geometric Representability:}
  Each element \( S_i \in S \) can be embedded into a Riemannian manifold \( \mathcal{M}_\kappa^d \) with constant curvature \( \kappa \in \mathbb{R} \), such that the embedding preserves its structural geometry (e.g., positive, negative or zero curvature).
\end{itemize}
Then, the set $ S $ is defined as \emph{geometric vocabulary} for graph structures.

\end{definition}

In constant curvature spaces~\cite{bachmann2020constant}, trees are modeled in hyperbolic space, cycles in spherical space and sequences in Euclidean space. This modeling paradigm effectively bridges topological heterogeneity~\cite{guo2025graphmore,he2025position}.
Based on the definition of geometric vocabulary, we leverage simple trees, cycles and sequences as the core structural patterns (detailed in Appendix \ref{C}).

\textbf{Overall Architecture.}
This work proposes the Riemannian graph tokenizer. The overall architecture is illustrated in Figure~\ref{fig:architecture}.
First, geometric vocabulary is sampled from the input graph and  structural coordinates are encoded in constant curvature space via Riemannian Encoder (Fréchet mean method)~\cite{DBLP:conf/www/ZhangWSLS21}. A novel Riemannian quantizer is designed to discretize the structural features and learn the structural codebooks. The quantized tokens are subsequently parallel transported or projected to the tangent space at the north pole to enable unified geometric fusion and decoding. Finally, the entire model is trained using a joint objective that combines reconstruction loss and contrastive loss.




\subsection{Riemannian Graph Tokenizer}\label{3.1}

RGT consists of three core components: the geometric vocabulary encoder, the Riemannian quantizer and the geometry-aligned decoder. We describe each of these modules in detail as follows.
\subsubsection{Geometric Vocabulary Encoder} \label{3.1.1}

Given a graph $\mathcal{G} = (\mathcal{V}, \mathcal{E})$, the Laplacian matrix is $L \in \mathbb{R}^{n \times n}$. We perform eigenvalue decomposition: $L = U \Lambda U^\top$, where $U = [\boldsymbol {u}_1, \boldsymbol {u}_2, \dots, \boldsymbol {u}_n]$ is the orthogonal matrix of eigenvectors, the $\Lambda = \operatorname{diag}(\lambda_1, \lambda_2, \dots, \lambda_n)$ is the diagonal matrix of eigenvalues. We select the eigenvectors corresponding to the top $d$ eigenvalues with the smallest magnitudes to construct a structural feature matrix $\boldsymbol X$, where each row of $\boldsymbol x$ serves as the initial structural feature of a node. 

The structural feature $\boldsymbol x$ is projected via a linear transformation $\psi$ to obtain $\boldsymbol x^{\mathcal{R}}$ in Euclidean space. For the hyperbolic and spherical spaces, we apply exponential mapping to project $\boldsymbol x^{\mathcal{R}}$ into the corresponding manifold as the initial coordinate. Therefore, the coordinate of a node can be uniformly denoted as $\boldsymbol x^{\mathcal{M}}, \mathcal{M} \in \{\mathcal{R},\mathcal{H},\mathcal{S}\}$ and its feature in the tangent space at the current point is denoted as $\boldsymbol e^{\mathcal{M}}$. We adopt the BFS algorithm~\cite{cormen2022introduction} to sample geometric vocabulary (trees, cycles and sequences) from the graph. The coordinates of these structural features are then updated via $\boldsymbol z^{\mathcal{M}} = \textbf{E}(\boldsymbol x^{\mathcal{M}})$, where $\boldsymbol x^{\mathcal{M}}, \boldsymbol z^{\mathcal{M}} \in \mathcal{M}$, $\textbf{E}$ denotes the geometric vocabulary encoder. The encoding procedure is detailed as follows:

For trees and cycles, we adopt the Fréchet Mean to update the coordinates of geometric vocabulary~\cite{lou2020differentiating,DBLP:conf/www/0008HZW0Y25}. Specifically, attention weights are computed by cross-geometry graph attention mechanism defined in Eq.(\ref{eq2}). The coordinate of each node is aggregated via the Fréchet mean on the corresponding manifold. The encoded coordinate is given by the geometric center $\boldsymbol{z}_i^{\mathcal{M}}, \mathcal{M} \in \{\mathcal{H}, \mathcal{S}\}$.

\begin{equation}\label{eq2}
     \alpha_{ij} = \frac{\exp\left(\phi\left([\boldsymbol{r}_i \, \| \, \boldsymbol{k}_j]\right)\right)}{\sum_{(i,t) \in S_i} \exp\left(\phi\left([\boldsymbol{r}_i \, \| \, \boldsymbol{k}_t]\right)\right)},
\end{equation}

\begin{equation}\label{eq3}
    \boldsymbol{z}_i^{\mathcal{M}} = \operatorname{mid}_\kappa\left(\left\{\boldsymbol{z}_j^{\mathcal{M}}, \alpha_{ij}\right\}_{(i,j) \in S_i}\right), 
\end{equation}
where $j$ denotes a descendant node of node $i$, $\phi$ is an arbitrary scalar-valued function and $\operatorname{mid}_\kappa(\cdot)$ denotes the Fréchet mean on the manifold. In the cross-geometry attention mechanism, the key, query and value are computed through manifold-preserving linear operations $f_*$~\cite{DBLP:conf/www/0008HZW0Y25} are denoted as:
\begin{equation}\label{eq4}
\boldsymbol{k}_i = f_{\boldsymbol{V}}(\boldsymbol{z}_i^{\mathcal{H}}), \quad
\boldsymbol{r}_i = f_{\boldsymbol{Q}}(\boldsymbol{z}_i^{\mathcal{S}}), \quad
\boldsymbol{v}_i = f_{\boldsymbol{V}}(\boldsymbol{z}_i^{\mathcal{H}}),   
\end{equation}
the coordinate $\boldsymbol z_i^\mathcal{M}$ of a node in the manifold space is obtained via Eq.(\ref{eq3}).


\paragraph{Geometric Center in Euclidean Space.}
For a sequence of length $L$, let the structural features of node $v_j$ be denoted as $\{\boldsymbol{z}_j^\mathcal{R}\}_{j=1}^L \subset \mathbb{R}^d$, with corresponding attention weights $\{w_{ij}\}_{j=1}^L$ satisfying $w_{ij} \ge 0$ and $\sum_{j=1}^L w_{ij} = 1$. Eq.(\ref{eq2}) is used to compute $w_{ij}$
within a single geometric space. The weighted geometric center (Fréchet Mean) in Euclidean space is then defined as:
\begin{equation}\label{eq5}
\boldsymbol{z_i^\mathcal{R}} = \frac{ \sum_{j=1}^{L} w_{ij} \boldsymbol{z}_j^\mathcal{R} }{ \left\| \sum_{j=1}^{L} w_{ij} \boldsymbol{z}_j^\mathcal{R} \right\|_2 }.
\end{equation}

As a result, we obtain the coordinates of the graph in three different spaces: $\boldsymbol z^\mathcal{H},\boldsymbol z^\mathcal{S}$ and $\boldsymbol z^\mathcal{R}$.



\subsubsection{Riemannian Quantizer}

\paragraph{Riemannian Quantizer.}
To enhance the model's ability to transfer structural knowledge across diverse graph domains, we design Riemannian Quantizer (RQ) that discretizes geometric vocabulary at the structural level. Specifically, for each constant curvature space $ \mathcal{M} \in \{\mathcal{R}, \mathcal{H}, \mathcal{S} \} $, we initialize the codebook $ \mathcal{C}^{\mathcal{M}}$ with $m$ codes, where each code vector corresponds to a point on the manifold. For the $i$-th node, its coordinate in space $\mathcal{M}$ is denoted as $\boldsymbol z_i^{\mathcal{M}}$. Following Eq.(\ref{eq6}), we select the code $ \boldsymbol q_i^{\mathcal{M}} \in \mathcal{C}^{\mathcal{M}} $ from the codebook that has the smallest geodesic distance to $\boldsymbol z_i^{\mathcal{M}} $. The core procedure is as follows:
\begin{equation}\label{eq6}
\mathrm{RQ:}
\left\{
\begin{aligned}
& j = \arg\min_jd_\mathcal{M}(\boldsymbol z_i^{\mathcal{M}},\boldsymbol c_j^{\mathcal{M}} ),\\
&\boldsymbol q_i^{\mathcal{M}}=\boldsymbol c_j^{\mathcal{M}},
\end{aligned}
\right.
\end{equation}
where $d_\mathcal{M}(\boldsymbol z_i^{\mathcal{M}},\boldsymbol c_j^{\mathcal{M}} )$ is geodesic distance. $\boldsymbol c_j^{\mathcal{M}} \in C^{\mathcal{M}}$ denotes the $j$-th code in the codebook. The nearest code is selected as $\boldsymbol q_i^{\mathcal{M}} = \boldsymbol c_j^{\mathcal{M}}$. 


To provide a theoretical foundation for the feasibility of RQ, we present two theorems: Cross-domain Encoding Consistency and Optimality of Riemannian Quantization. The details are as follows:

\begin{theorem}[Cross-domain Encoding Consistency]
Let $ z_i^{\mathcal{M}} \in D_1$, $z_j^{\mathcal{M}} \in D_2 $, where $D_1, D_2 \in \mathcal{D}$, if $ d(z_i^{\mathcal{M}}, z_j^{\mathcal{M}}) \le \epsilon $, then the following holds: 
$$ d(RQ(z_i^{\mathcal{M}}), RQ(z_j^{\mathcal{M}})) \le 3 \epsilon. $$
\end{theorem}

The theorem 1 establishes that RQ preserves local consistency, we now show that RQ also enjoys a global optimality on Riemannian manifolds.

\begin{theorem}[Optimality of Riemannian Quantization]
Let $(\mathcal{M}, g)$ be a smooth Riemannian manifold equipped with a geodesic distance function $d_{\mathcal{M}}(\cdot,\cdot)$. Given a set of geometric vocabulary $\{\mathcal{S}_i^{(j)}\}_{j=1}^M$ and a structural codebook $\mathcal{C} = \{\boldsymbol{c}_j^{\mathcal{M}}\}_{j=1}^m \subset \mathcal{M}$, the quantizer $RQ:\{\mathcal{S}_i^{(j)}\}_{j=1}^M \to \mathcal{C}$ is defined as:
\begin{equation}\label{eq7}
\boldsymbol{c}_j^{\mathcal{M}} = \arg\min_{\boldsymbol{c} \in \mathcal{C}} \mathbb{E}\left[ d_{\mathcal{M}}( \boldsymbol{z}, \boldsymbol{c} ) \,\middle|\, \boldsymbol{z} \in \{\mathcal{S}_i^{(j)}\}_{j=1}^M \right],
\end{equation}
the quantizer $RQ$ minimizes the overall expected geodesic distortion $\mathbb{E}[ d_{\mathcal{M}}( \boldsymbol{z}, RQ(\boldsymbol{z}) ) ]$, where $\boldsymbol{z} \sim P(\{\mathcal{S}_i^{(j)}\}_{j=1}^M)$ denotes the probability distribution over structural tokens within the geometric vocabulary set $\{\mathcal{S}_i^{(j)}\}_{j=1}^M$.

\end{theorem}

\paragraph{Geometric Interpretation 1.}

The theorem 1 establishes that RQ preserves the neighborhood relations of geometric vocabularies across different domains at the local level, thereby ensuring that structurally similar patterns remain close after quantization. Furthermore, the theorem 2 provides a global guarantee by showing that selecting the codeword for each geometric vocabulary $\{\mathcal{S}_i^{(j)}\}_{j=1}^M$ as its geodesic center on the manifold $\mathcal{M}$ minimizes the quantization error. 
Together, these theoretically validate that the RQ formulation in Eq.~(\ref{eq6}) offers an optimal structure-level discretization strategy: structurally similar geometric vocabularies (such as trees, cycles, and sequences) can be consistently mapped to the same code in the corresponding codebook, thereby enabling cross-domain structural alignment. 
The detailed proofs are provided in Appendix~\ref{C}.

\paragraph{Riemannian Straight-Through Estimator.}
To maintain the intrinsic structure of the manifold while supporting end-to-end differentiable training, we propose the Riemannian Straight-Through Estimator (RSTE), formulated as Eq.(\ref{eq8}):
\begin{equation}\label{eq8}
\hat{\boldsymbol z_i}^{\mathcal{M}} = \exp_{\boldsymbol z_i^{\mathcal{M}}}\left( \mathrm{sg}\left[ \log_{\boldsymbol z_i^{\mathcal{M}}}(\boldsymbol q_i^{\mathcal{M}}) \right] \right), \quad \boldsymbol q_i^{\mathcal{M}} \in \mathcal{M},
\end{equation}
where the operator $\log_{\boldsymbol z_i^{\mathcal{M}}}(\boldsymbol q_i^{\mathcal{M}})$ denotes the logarithmic map that projects the point $\boldsymbol q_i^{\mathcal{M}}$ onto the tangent space at $\boldsymbol z_i^{\mathcal{M}}$, while $\exp_{\boldsymbol z_i^{\mathcal{M}}}(\boldsymbol \cdot)$ denotes the exponential map that lifts a tangent vector back onto the manifold. $\mathrm{sg}[\cdot]$ represents the stop-gradient operator.

\paragraph{Geometric Interpretation 2.} In Euclidean space, vector quantizer~\cite{van2017neural} typically adopts $\boldsymbol z^{\mathcal{R}} = \boldsymbol z^{\mathcal{R}} + \mathrm{sg}[\boldsymbol q^{\mathcal{R}} - \boldsymbol z^{\mathcal{R}}]$ as a Straight-Through Estimator (STE)~\cite{bengio2013estimating}, where $\boldsymbol q^{\mathcal{R}}$ is treated as a constant during backpropagation and does not receive gradient updates. However, this estimator fails to preserve the intrinsic structure of non-Euclidean manifolds. We propose the RSTE to address this limitation. The advantage of this method lies in the fact that the quantization error $\log_{z_i}^{\mathcal{M}}(q_i^{\mathcal{M}})$ is retained in the tangent space, where the geometric structure is more stable. Since gradients are propagated within the tangent space, the manifold $\mathcal{M}$ maintains geometric consistency. It prevents the distortion typically introduced by Euclidean additive operations. This effectively alleviates the issue of codebook collapse~\cite{zhu2024addressing,fifty2024restructuring}. The multiple similar geometric vocabularies are aligned to the same code in the codebook. Quantization in Riemannian space facilitates structural knowledge transfer. Detailed analyses are provided in Appendix~\ref{D}. 

\subsubsection{Geometry-Aligned Decoder}\label{3.1.3}
As the quantized tokens reside in different geometric spaces, we propose a geometry-aligned decoder $\hat{\boldsymbol x}_i^{\mathcal{M}} = \mathrm{D}(\boldsymbol q_i^{\mathcal{M}})$ to obtain consistent representation and ensure spatially aligned reconstruction and contrast. For the Euclidean-space token $\boldsymbol q_i^\mathcal{R}$, we first project it to the tangent space at the north pole to obtain $\boldsymbol{e}_i^{\mathcal{R}}$. Then, for hyperbolic and spherical spaces, the Euclidean feature $\boldsymbol q_i^\mathcal{R}$ is projected into the tangent space at $\boldsymbol q_i^\mathcal{M} (\mathcal{M} \in {\mathcal{H}, \mathcal{S}})$, resulting in $\boldsymbol {e}_i^{\mathcal{M}}$. Riemannian parallel transport is utilized to align $\boldsymbol{e}_i^{\mathcal{M}}$ from the point $\boldsymbol q_i^\mathcal{M} (\mathcal{M} \in {\mathcal{H}, \mathcal{S}})$ to the tangent space $T_{\boldsymbol{o_k}}\mathcal{M}$ at the north pole. The aligned representation in the tangent space, $\boldsymbol{e}_i^{T_{\boldsymbol{o_k}}\mathcal{M}}$, is defined as follows:

\begin{equation}\label{eq9}
\boldsymbol{e}_i^{T_{\boldsymbol{o_k}}\mathcal{M}} = PT_{\boldsymbol q_i^\mathcal{M} \rightarrow \boldsymbol{o_k}}\left( \boldsymbol{e}_i^{\mathcal{M}} \right), 
\end{equation}
where $PT_{\boldsymbol q_i^\mathcal{M} \rightarrow \boldsymbol{o_k}}(\cdot)$ denotes the parallel transport operation from a point $\boldsymbol q_i^\mathcal{M}$ on the manifold to the north pole $\boldsymbol{o_k}$. The resulting embeddings in the tangent space at the north pole are denoted as $\boldsymbol{e}_i^{T_{\boldsymbol{o_k}}\mathcal{H}}$ and $\boldsymbol{e}_i^{T_{\boldsymbol{o_k}}\mathcal{S}}$ for the hyperbolic and spherical spaces, respectively. Next, the aligned embeddings from different manifolds are concatenated and projected back to the original space. Geometry-aligned representations from all spaces are fused into a single unified representation:

\begin{equation}\label{eq10}
\hat{x}_i = \textbf{Fuse} \left( \mathbf{e}_i^{\mathcal{R}}, \boldsymbol{e}_i^{T_{\boldsymbol{o_k}}\mathcal{H}}, \boldsymbol{e}_i^{T_{\boldsymbol{o_k}}\mathcal{S}} \right),
\end{equation}
where $\textbf{Fuse}(\cdot)$ denotes a simple fusion operation, such as concatenation followed by a linear transformation, and $\hat{\boldsymbol{x}}_i$ represents the reconstructed structural feature. The geometry-aligned decoder enables unified decoding of embeddings across geometric spaces with varying curvatures, ensuring consistency in representation.

\subsection{Loss Function and Optimization}

\paragraph{Loss Function.} We design a unified loss function to jointly optimize the Riemannian graph tokenizer across constant curvature spaces. Each geometric space maintains an independent codebook. The loss function integrates both reconstruction loss $\mathcal{L}_{rec} $ and geometric contrastive loss $ \mathcal{L}_{cl}$. 

\begin{equation}\label{eq11}
\mathcal{L} =\epsilon \mathcal{L}_{rec} + \lambda \mathcal{L}_{cl},
\end{equation}

The reconstruction loss and geometric contrastive loss are formulated in Eq.(\ref{eq12}) and Eq.(\ref{eq13}):
\begin{equation}\label{eq12}
\begin{aligned}
\mathcal{L}_{rec} 
&= \frac{1}{n} \sum_{i=1}^n \| \hat{\boldsymbol x}_i - \boldsymbol x_i \|^2 + \frac{1}{n} \sum_{\mathcal{M}} \Bigg[
    \beta \sum_{i=1}^n d_\mathcal{M}^2(\boldsymbol z_i^\mathcal{M}, \mathrm{sg}(\boldsymbol q_i^\mathcal{M})) \\
& + \gamma \sum_{i=1}^n d_\mathcal{M}^2(\mathrm{sg}(\boldsymbol z_i^\mathcal{M}), \boldsymbol q_i^\mathcal{M})
\Bigg],
\end{aligned}
\end{equation}

\begin{equation}\label{eq13}
\begin{aligned}
& \mathcal{L}_{cl} = \underbrace{
    \left[ \mathcal{J}(\mathcal{R}, \mathcal{H}) + \mathcal{J}(\mathcal{R}, \mathcal{S}) + \mathcal{J}(\mathcal{H}, \mathcal{S}) \right]
}_{\text{Cross-geometry contrastive loss}},
\end{aligned}
\end{equation}

\begin{equation}\label{eq14}
\resizebox{0.9\linewidth}{!}{$
\begin{aligned}
& \mathcal{J}(\mathcal{H}, \mathcal{S})=-\sum_{i=1}^N \log \frac{\exp \left(\left\langle P T_{\boldsymbol q_i^{\mathcal{H}} \rightarrow \boldsymbol o_k}\left(\boldsymbol e_i^\mathcal{H}\right), P T_{q_i^\mathcal{S} \rightarrow \boldsymbol o_k}\left(\boldsymbol e_i^\mathcal{S}\right)\right\rangle\right)}{\sum_{j=1}^N \exp \left(\left\langle P T_{\boldsymbol q_i^\mathcal{H}\rightarrow \boldsymbol o_k}\left(\boldsymbol e_i^\mathcal{H}\right), P T_{\boldsymbol q_j^\mathcal{S} \rightarrow o_k}\left(\boldsymbol e_j^\mathcal{S}\right)\right\rangle\right)},
\end{aligned}$}
\end{equation}
where the  ${\boldsymbol {q}}_i^\mathcal{M}$ denotes the representation of the $i$-th node in the geometric space $\mathcal{M}$. After parallel transport, it is projected into the shared tangent space to obtain the fused embedding $\hat{\boldsymbol {x}}_i$. The term $\mathcal{J}(H, S)$ in Eq.(\ref{eq13}) denotes the contrastive learning loss computed in the shared tangent space, encouraging alignment between representations from different geometric spaces. The other two follow similarly. The reconstruction loss is computed by comparing the reconstructed representation $\hat{\boldsymbol x}_i$ with the original input $\boldsymbol x_i$.

\paragraph{Optimization}
We update each variable. First, Eq.(\ref{eq15}) is the gradient propagation for the $\boldsymbol z_i^{\mathcal{M}}$:

\begin{equation}\label{eq15}
\begin{aligned}
\boldsymbol{z}_i^{\mathcal{M}, (t+1)} =
& \boldsymbol{z}_i^{\mathcal{M}, (t)} - \eta  \Bigg[ \frac{\partial \mathcal{L}}{\partial \hat{\boldsymbol{x}}_i}
\cdot \frac{\partial \hat{\boldsymbol{x}}_i}{\partial \hat{\boldsymbol{z}}_i^{\mathcal{M}}}
\cdot \frac{\partial \exp_{\boldsymbol{z}_i^{\mathcal{M}}}^{\mathcal{M}} 
\left( \mathrm{sg}\left[ \log_{\boldsymbol{z}_i^{\mathcal{M}}}(\boldsymbol{q}_i^{\mathcal{M}})\right] \right)}{\partial \boldsymbol{z}_i^{\mathcal{M}}} \\
& + \beta \cdot \frac{\partial}{\partial \boldsymbol{z}_i^{\mathcal{M}}} d_\mathcal{M}^2(\boldsymbol{z}_i^\mathcal{M}, \mathrm{sg}( \boldsymbol{q}_i^\mathcal{M})) 
+ \lambda \cdot \frac{\partial \mathcal{L}_{cl}}{\partial \boldsymbol{z}_i^{\mathcal{M}}}
\Bigg].
\end{aligned}
\end{equation}

Eq.(\ref{eq16}) is the gradient propagation for the code $\boldsymbol q_i^{\mathcal{M}}$:
\begin{equation}\label{eq16}
\boldsymbol q_i^{\mathcal{M},(t+1)} = \boldsymbol q_i^{\mathcal{M},(t)} - \eta(\gamma\frac{\partial}{\partial \boldsymbol q_i^\mathcal{M}} d_{\mathcal{M}}^2(\mathrm{sg}(\boldsymbol z_i^\mathcal{M}), \boldsymbol q_i^\mathcal{M})+ \lambda\frac{\partial \mathcal{L}_{cl}}{\partial {\boldsymbol z}_i^{\mathcal{M}}}),
\end{equation}
where the $t$ is the iteration. The above design ensures geometric consistency between structural tokens and discrete codes, with the training process strictly following curvature-aware geodesic paths. This enhances the model's stability and performance in structural transfer and cross-domain generalization.

The proposed model is pretrained using the combined loss Eq.(\ref{eq11}). The overall training process is summarized in Algorithm 1 (Appendix~\ref{E}), with a computational complexity of $\mathcal{O}(|\mathcal{V}|^2 + |\mathcal{E}|)$, where $|\mathcal{V}|$ and $|\mathcal{E}|$ denote the number of nodes and edges, respectively. The Riemannian graph tokenizer effectively captures globally consistent structural patterns across diverse domains.
\vspace{-0.5em}
\section{Experiment}\label{sec4}

\begin{table*}[htbp]
  \caption{Cross-domain transfer learning performance of node classification on four datasets.}
  \vspace{-10pt}
  \label{tab1}
  \centering
  \LARGE
  \renewcommand{\arraystretch}{1}
  \resizebox{0.87\textwidth}{!}{
  \begin{tabular}{c| cc  cc  cc cc}
  \hline
  & \multicolumn{8}{c}{\textbf{Node Classification Results}}  \\ 
  \cline{2-9}
  \textbf{Method}  
  & \multicolumn{2}{c}{\textbf{Pubmed}} 
  & \multicolumn{2}{c}{\textbf{Github}} 
  & \multicolumn{2}{c}{\textbf{USA}}
  & \multicolumn{2}{c}{\textbf{Europe}} \\
  \cline{2-9}
  & ACC & F1 
  & ACC & F1 
  & ACC & F1 
  & ACC & F1\\
  \hline
  DGI 
  & 75.66{$\pm$0.86} 
  & 73.66{$\pm$1.46}
  & 74.10{$\pm$0.15} 
  & 42.56{$\pm$0.05} 
  & 38.90{$\pm$1.57}
  & 33.06{$\pm$3.98}
  & 24.48{$\pm$0.97}
  & 9.83{$\pm$0.31}\\
  GraphMAE 
  & \textbf{79.47{$\pm$0.45} }
  & \textbf{78.75{$\pm$0.40} }
  & 67.57{$\pm$2.76} 
  & 63.14{$\pm$1.81}
  & 32.78{$\pm$6.63}
  & 21.42{$\pm$5.39}
  & 31.28{$\pm$2.10} 
  & 20.96{$\pm$3.47}\\
  OFA
  & 52.23{$\pm$6.45} 
  & 50.35{$\pm$3.34}
  & - 
  & - 
  & - 
  & - 
  & - 
  & -  \\
  ZeroG
  & 69.12{$\pm$0.89} 
  & 68.35{$\pm$1.23}
  & - & - & - & - & - & - \\
  Graphany
  & 74.18{$\pm$2.57} 
  & 70.87{$\pm$0.18} 
  & 79.26{$\pm$0.20}
  & 76.93{$\pm$0.42}
  & 45.73{$\pm$1.70}
  & 45.35{$\pm$0.39}
  & 34.54{$\pm$1.57}
  & 27.54{$\pm$0.42}\\
  OpenGraph 
  & 68.11{$\pm$1.98} 
  & 65.45{$\pm$3.33} 
  & 47.85{$\pm$0.54} 
  & 46.34{$\pm$0.78}
  & 40.65{$\pm$1.38}
  & 42.89{$\pm$1.14}
  & 27.36{$\pm$1.63}
  & 25.74{$\pm$0.73}\\
  GFT
  & 70.30{$\pm$1.76} 
  & 70.38{$\pm$1.83}
  & - & - & - & - & - & - \\
  RiemannGFM
  & 73.94{$\pm$5.66} 
  & 74.25{$\pm$4.25} 
  & 85.51{$\pm$0.39} 
  & 85.24{$\pm$0.15}
  & 55.63{$\pm$2.33}
  & 55.92{$\pm$2.22}
  & 36.50{$\pm$4.68}
  & 30.77{$\pm$4.21}\\
  \textbf{RGT}(Ours)
  & 76.22{$\pm$0.80} 
  & 76.14{$\pm$0.77}
  & \textbf{86.35{$\pm$0.14}}
  & \textbf{85.67{$\pm$0.24}}
  & \textbf{56.72{$\pm$2.51} }
  & \textbf{55.80{$\pm$3.29}}
  & \textbf{37.75{$\pm$3.39}}
  & \textbf{32.17{$\pm$1.48}}\\
    \hline
    \end{tabular}}
    \vspace{-5pt}
    \end{table*}

\begin{table*}[htbp]
  \caption{Cross-domain transfer learning performance of link prediction on four datasets.}
  \vspace{-5pt}
  \label{tab2}
  \centering
  \LARGE
  \renewcommand{\arraystretch}{1}
  \resizebox{0.87\textwidth}{!}{
  \begin{tabular}{c| cc  cc  cc cc}
  \hline
  &\multicolumn{8}{c}{\textbf{Link Prediction Results}}  \\ 
  \cline{2-9}
  \textbf{Method} 
  & \multicolumn{2}{c}{\textbf{Pubmed}} 
  & \multicolumn{2}{c}{\textbf{Github}} 
  & \multicolumn{2}{c}{\textbf{USA}}
  & \multicolumn{2}{c}{\textbf{Europe}} \\
  \cline{2-9}
  & AUC & AP 
  & AUC & AP  
  & AUC & AP  
  & AUC & AP  \\
  \hline
  DGI 
  & 56.19{$\pm$0.19} 
  & 61.56{$\pm$0.35}
  & 55.75{$\pm$8.68}
  & 53.66{$\pm$7.79}
  & 56.17{$\pm$2.31}
  & 48.39{$\pm$1.35}
  & 58.82{$\pm$5.38}
  & 51.38{$\pm$3.30}\\
  GraphMAE
  & 84.40{$\pm$0.38}
  & 82.36{$\pm$0.36}
  & 65.94{$\pm$0.18}
  & 67.51{$\pm$0.14}
  & 65.81{$\pm$6.51}
  & 67.36{$\pm$1.27}
  & 45.58{$\pm$1.73}
  & 46.28{$\pm$0.81}\\
  OFA
  & 85.32{$\pm$1.67} 
  & 83.28{$\pm$1.39}
  & - & - & - & - & - & - \\
  OpenGraph 
  & 65.04{$\pm$0.98}
  & 64.31{$\pm$3.98} 
  & 78.34{$\pm$0.92}
  & 79.76{$\pm$1.20}
  & 77.89{$\pm$2.34}
  & 79.32{$\pm$1.32}
  & 72.34{$\pm$1.51}
  & 70.32{$\pm$1.22}\\
  RiemannGFM
  & 89.84{$\pm$0.32} 
  & 88.99{$\pm$0.27}
  & 88.49{$\pm$0.22}
  & 87.78{$\pm$0.17}
  & \textbf{93.57{$\pm$0.23}}
  & \textbf{96.02{$\pm$0.16}}
  & \textbf{85.49{$\pm$0.98}}
  & \textbf{82.06{$\pm$0.80}}\\
  \textbf{RGT}(Ours)
  & \textbf{93.28{$\pm$0.01}}
  & \textbf{91.92{$\pm$0.01} }
  & \textbf{90.31{$\pm$0.03}}
  & \textbf{88.90{$\pm$0.02}}
  & 93.32{$\pm$0.03}
  & 92.89{$\pm$0.03}
  & 85.17{$\pm$0.07}
  & 82.84{$\pm$0.07}\\
  \hline
  \end{tabular}}
  
\end{table*}

In this section, we aim to answer the following research questions:
\textbf{RQ1.} How does RGT perform in cross-domain transfer learning?
\textbf{RQ2.} How effective is RGT under few-shot learning scenarios?
\textbf{RQ3.} How important is the use of geometric vocabulary in RGT?
\textbf{RQ4.} How does the choice of pretraining dataset influence the performance of RGT? 

\noindent \textbf{RQ5.} How does RGT scale to web-scale graphs? Additional experiments are provided in Appendix~\ref{F}. \textbf{Codes} are available at {\color{blue}\url{https://github.com/Miyla77/RGT.git}}.

\subsection{Experimental Setups}\label{4.1}

\indent \textbf{Hyperparameter settings:} The number of layers for tree-based sampling is set to 2, cycles are constructed as triangles (3-node), and the sequence length is set to $L$. The embedding dimension $d$ for structural embedding is 32. The curvatures of the constant curvature spaces $\kappa$ are set to 1 (spherical), $-1$ (hyperbolic) and 0 (Euclidean space). The feature dimension of each code is 32, the number of code in codebook $m=128$. All hyperparameters are selected via grid search. For baselines, the hyperparameters strictly follow those reported in the original papers.

\textbf{Pretraining and Evaluation:} The pretraining datasets used for our model is the same as in RiemannGFM. During evaluation, the pretrained codebooks are kept frozen while the GNN classifier is fine-tuned on the target test datasets. Other models are evaluated on the same test datesets. The results are averaged over multiple runs to reduce the impact of randomness. 

\textbf{Datasets.} We conduct experiments using several graph datasets, which are categorized based on their attribute types and application contexts. Specifically, we select three datasets ogbn-arxiv~\cite{nips2020arxiv}, Computers and Physics~\cite{physics_computers} for the pretraining phase. For the main evaluation experiments, we employ seven datasets: three text-attributed graphs (Cora, PubMed and Citeseer)~\cite{kipf2016semi}, one mix-attributed graph (GitHub)~\cite{DBLP:journals/compnet/RozemberczkiAS21}, three non-attributed graphs (Airport of USA, Europe and Brazil)~\cite{DBLP:conf/kdd/RibeiroSF17} and one Web-scale graph ogbn-products~\cite{hu2020open}.

\textbf{Baselines \& Metrics.} We choose seven competitive baselines, categorized into two groups. Self‑supervised models: DGI~\cite{DBLP:conf/iclr/VelickovicFHLBH19} and GraphMAE2~\cite{hou2023graphmae2}. Graph foundation models: OFA~\cite{DBLP:conf/iclr/0057FKLT0Z24}, ZeroG~\cite{DBLP:conf/kdd/0001WLY024}, GraphAny~\cite{zhaofully}, OpenGraph~\cite{DBLP:conf/emnlp/XiaK024}, GFT~\cite{DBLP:conf/nips/WangZCZ024}, RiemannGFM~\cite{DBLP:conf/www/0008HZW0Y25}.
Experiments cover node‑classification and link‑prediction settings. For node classification we report overall accuracy (ACC) together with Micro‑F1 (F1). For link prediction we measure area under the AUC and average precision (AP). We report the mean and standard deviation over 5 seeds for each result. Best results are bolded. The dash means the model cannot handle non-attributed graphs.




\subsection{Results and Discussion}
\paragraph{Cross-domain Transfer Learning Performance (RQ1).}

The experimental results of node classification and link prediction are shown in Table~\ref{tab1} and Table~\ref{tab2}, respectively. RGT demonstrates significant advantages. On citation networks dataset, RGT achieves comparable node classification performance to traditional self-supervised methods that are trained and evaluated within the same domain. Due to the incorporation of structural vocabulary, RGT yields superior results in link prediction. Meanwhile, RGT achieves better performance than existing GFMs on both node classification and link prediction tasks. This holds true even when models like GFT utilize multi-domain quantization on text-attributed graphs and OpenGraph incorporates graph tokenizer. The major strength of RGT lies in its ability to handle non-attributed graphs by structural features, where it outperforms RiemannGFM. These findings validate the effectiveness of structure-level quantization and the cross-domain transferability enabled by RGT.

\begin{table*}[ht]
  \caption{One-shot learning performance of node classification on four datasets.}
  \vspace{-10pt}
  \label{tab3}
  \centering
  \LARGE
  \renewcommand{\arraystretch}{1}
  \resizebox{0.87\textwidth}{!}{
  \begin{tabular}{c| cc  cc  cc cc}
  \hline
  &\multicolumn{8}{c}{\textbf{Node Classification Results}}  \\ 
  \cline{2-9}
  \textbf{Method} 
  & \multicolumn{2}{c}{\textbf{Pubmed}} 
  & \multicolumn{2}{c}{\textbf{Github}} 
  & \multicolumn{2}{c}{\textbf{USA}}
  & \multicolumn{2}{c}{\textbf{Europe}} \\
  \cline{2-9}
  & ACC & F1 
  & ACC & F1 
  & ACC & F1
  & ACC & F1\\
  \hline
  DGI
  & 54.09{$\pm$1.92} 
  & 45.66{$\pm$7.34}
  &44.26{$\pm$0.14} 
  & 44.72{$\pm$3.03}
  & 26.69{$\pm$2.94}
  & 13.58{$\pm$4.84}
  & 25.62{$\pm$0.00}
  & 10.20{$\pm$0.00}\\
  GraphMAE
  & 47.03{$\pm$7.27} 
  & 42.31{$\pm$9.40}
  & 55.25{$\pm$2.03} 
  & 46.78{$\pm$1.74}
  & 23.69{$\pm$1.70}
  & 14.73{$\pm$2.83}
  & 27.07{$\pm$3.20}
  & 18.54{$\pm$2.30}\\
  OFA
  & 41.25{$\pm$0.94} 
  & 38.45{$\pm$2.96}
  & - & - & - & - & - & - \\
  ZeroG
  & \textbf{74.52{$\pm$0.52} }
  & \textbf{74.15{$\pm$0.15}}
  & - & - & - & - & - & - \\
  Graphany
  & 72.46{$\pm$6.89} 
  & 70.55{$\pm$4.27} 
  & 63.69{$\pm$6.39} 
  & 43.67{$\pm$0.47}
  & 26.64{$\pm$0.06}
  & 27.00{$\pm$0.32}
  & 26.57{$\pm$0.57}
  & 23.01{$\pm$4.37}\\
  OpenGraph 
  & 66.96{$\pm$5.34} 
  & 63.86{$\pm$4.88}
  & 32.67{$\pm$0.12} 
  & 45.23{$\pm$1.21}
  & 30.45{$\pm$3.71}
  & 29.32{$\pm$2.46}
  & 22.46{$\pm$3.89}
  & 20.65{$\pm$3.22}\\
  GFT
  & 73.56{$\pm$3.15} 
  & 74.55{$\pm$4.70}
  & - & - & - & - & - & -\\
  RiemannGFM
  & 44.92{$\pm$4.28}
  & 37.28{$\pm$3.54} 
  & 64.32{$\pm$5.25}
  & 63.96{$\pm$4.94}
  & 34.34{$\pm$5.53}
  & 34.29{$\pm$5.55}
  & 27.58{$\pm$4.89}
  & 21.57{$\pm$5.27}\\
  \textbf{RGT}(Ours)
  & 52.18{$\pm$7.97} 
  & 47.45{$\pm$9.74}
  & \textbf{72.28{$\pm$6.74}}
  & \textbf{70.39{$\pm$8.00}} 
  & \textbf{36.13{$\pm$5.67}}
  & \textbf{33.34{$\pm$3.09}}
  & \textbf{28.75{$\pm$6.57}}
  & \textbf{21.94{$\pm$6.75}}\\
  \hline
  \end{tabular}
  }
  \vspace{-5pt}
\end{table*}

\begin{table*}[ht]
  \caption{Five-shot learning performance of node classification on four datasets.}
  \vspace{-5pt}
  \label{tab4}
  \centering
  \LARGE
  \renewcommand{\arraystretch}{1}
  \resizebox{0.87\textwidth}{!}{
  \begin{tabular}{c| cc  cc  cc cc}
  \hline
  &\multicolumn{8}{c}{\textbf{Node Classification Results}}  \\ 
  \cline{2-9}
  \textbf{Method} 
  & \multicolumn{2}{c}{\textbf{Pubmed}} 
  & \multicolumn{2}{c}{\textbf{Github}} 
  & \multicolumn{2}{c}{\textbf{USA}}
  & \multicolumn{2}{c}{\textbf{Europe}} \\
  \cline{2-9}
  & ACC & F1 
  & ACC & F1 
  & ACC & F1
  & ACC & F1\\
  \hline
  DGI
  & 64.07{$\pm$8.43} 
  & 61.88{$\pm$4.84}
  & 75.56{$\pm$1.37} 
  & 51.21{$\pm$7.40}
  & 35.84{$\pm$1.28}
  & 28.02{$\pm$3.20}
  & 25.62{$\pm$0.00}
  & 10.20{$\pm$0.00}\\
  GraphMAE
  & 72.09{$\pm$1.86} 
  & 71.62{$\pm$2.28}
  & 57.12{$\pm$2.16} 
  & 52.39{$\pm$1.97}
  & 31.42{$\pm$5.58}
  & 21.11{$\pm$4.59}
  & 26.74{$\pm$1.59}
  & 14.20{$\pm$3.00}\\
  OFA
  & 41.26{$\pm$4.36} 
  & 39.67{$\pm$5.35}
  & - & - & - & - & - & - \\
  ZeroG
  & 72.26{$\pm$3.21} 
  & 69.48{$\pm$2.39}
  & - & - & - & - & - & - \\
  Graphany 
  & 62.82{$\pm$6.16} 
  & 60.28{$\pm$4.15} 
  & 54.52{$\pm$10.07}
  & 44.66{$\pm$0.47}
  & 27.69{$\pm$0.08}
  & 28.93{$\pm$0.25}
  & 29.54{$\pm$4.24}
  & 23.21{$\pm$5.12}\\
  OpenGraph 
  & 65.62{$\pm$3.68} 
  & 62.78{$\pm$3.28}
  & 45.93{$\pm$0.92} 
  & 42.46{$\pm$1.21}
  & 40.23{$\pm$3.89}
  & 39.56{$\pm$2.92}
  & 32.21{$\pm$2.14}
  & 23.42{$\pm$3.22}\\
  GFT
  & \textbf{73.89{$\pm$2.76} }
  & \textbf{74.26{$\pm$4.39}}
  & - & - & - & - & - & -\\
  RiemannGFM
  & 64.54{$\pm$5.32} 
  & 63.84{$\pm$4.42}
  & 82.34{$\pm$2.46} 
  & 82.14{$\pm$2.15}
  & 37.41{$\pm$1.25}
  & 32.41{$\pm$1.31}
  & 30.18{$\pm$5.54}
  & 21.87{$\pm$5.14}\\
  \textbf{RGT}(Ours)
  & 70.90 {$\pm$5.20} 
  & 70.73 {$\pm$5.52}
  & \textbf{85.12{$\pm$2.12}}
  & \textbf{84.96{$\pm$1.98}}
  & \textbf{40.34{$\pm$3.24}}
  & \textbf{33.43{$\pm$3.15}}
  & \textbf{33.10{$\pm$3.62}}
  & \textbf{26.55{$\pm$2.97}}\\
  \hline
  \end{tabular}
  }
  \vspace{-10pt}
\end{table*}
\paragraph{Few-shot Learning Performance(RQ2).}
The results of one-shot and five-shot node classification are reported in Table~\ref{tab3} and Table~\ref{tab4}, respectively. RGT demonstrates strong transferability in this low-resource setting. On the Pubmed dataset, methods such as ZeroG, Graphany and GFT leverage test-domain data during pretraining. In contrast, RGT performs strict cross-domain transfer without any exposure to the target domain. Nevertheless, RGT still exhibits competitive overall results. On the Github dataset, RGT achieves competitive performance on both evaluation metrics, showcasing its strong capacity to model complex graph structures. In the five-shot setting, as more structural information becomes available, the performance of RGT improves significantly. Except for the GraphMAE and GFT, RGT outperforms other methods, particularly on non-attributed graphs, where its advantage becomes more evident. These experimental results suggest that RGT effectively leverages limited supervision to capture semantic correlations between graph structures and class labels, making it well-suited for graph learning tasks under low-label regimes.

\paragraph{Importance of Geometric Vocabulary(RQ3).}

Graphs can be regarded as discrete manifolds. Therefore, modeling trees in hyperbolic space, cycles in spherical space and sequences in Euclidean space aligns well with the mathematical formulation of Riemannian geometry. To verify the effectiveness of the proposed geometric vocabulary, we assign trees, cycles and sequences to different manifolds: $\mathcal{H}^{32}_{-1}$ (hyperbolic), $\mathcal{S}^{32}_{1}$ (spherical) and $\mathcal{R}^{32}_{0}$ (Euclidean), resulting in nine possible combinations. The results across various datasets are reported in Table~\ref{tab5}. The results show that assigning trees to hyperbolic space, cycles to spherical space and sequences to Euclidean space yields the best performance among all configurations. In contrast, combinations where all three vocabularies are embedded in mismatched spaces (Row 4 and Row 6 in the table) or one space (Row 7-9 in the table) lead to significantly worse results. These findings confirm the effectiveness of the geometry vocabulary in this work.
\begin{table}[htbp]
\centering
\caption{Link prediction results of geometric ablation are reported in terms of AUC.}
\vspace{-10pt}
\label{tab5}
\LARGE
\renewcommand{\arraystretch}{1}
\resizebox{\columnwidth}{!}{
\begin{tabular}{ccc|c c c c}
\hline
\textbf{Trees} & \textbf{Cycles} & \textbf{Sequences} & \textbf{Cora} & \textbf{Citeseer} & \textbf{Pubmed} & \textbf{USA} \\
\hline
$\mathcal H^{32}_{-1}$ & $\mathcal S^{32}_{1}$ & $\mathcal R^{32}_{0}$ 
& \textbf{80.72$\pm$0.97} & \textbf{66.40$\pm$1.53} & \textbf{76.26$\pm$0.73} & \textbf{56.72$\pm$2.51} \\
$\mathcal H^{32}_{-1}$ & $\mathcal R^{32}_{0}$ & $\mathcal S^{32}_{1}$ 
& 78.53$\pm$0.78 & 64.98 $\pm$ 2.22 & 74.52$\pm$0.81 & 52.69$\pm$1.47 \\
$\mathcal S^{32}_{1}$ & $\mathcal H^{32}_{-1}$ & $\mathcal R^{32}_{0}$ 
& 79.42$\pm$0.83 & 65.77$\pm$1.58 & 74.34$\pm$0.82 & 52.77$\pm$1.05 \\
$\mathcal S^{32}_{1}$ & $\mathcal R^{32}_{0}$ & $\mathcal H^{32}_{-1}$ 
& 77.98$\pm$1.21 & 64.58$\pm$2.56 & 73.88$\pm$0.89 & 51.13$\pm$1.35 \\
$\mathcal R^{32}_{0}$ & $\mathcal S^{32}_{1}$ & $\mathcal H^{32}_{-1}$ 
& 78.32$\pm$0.53 & 65.64$\pm$1.20 & 75.26$\pm$0.63 & 53.71$\pm$1.24 \\
$\mathcal R^{32}_{0}$ & $\mathcal H^{32}_{-1}$ & $\mathcal S^{32}_{1}$ 
& 76.85$\pm$1.21 & 63.69$\pm$1.32 & 74.26$\pm$0.93 & 52.21$\pm$1.37 \\
$\mathcal H^{32}_{-1}$ & $\mathcal H^{32}_{-1}$ & $\mathcal H^{32}_{-1}$ 
& 78.80$\pm$0.48 & 64.80$\pm$1.11 & 73.36$\pm$0.83  &  53.57$\pm$1.02 \\
$\mathcal S^{32}_{1}$ & $\mathcal S^{32}_{1}$ & $\mathcal S^{32}_{1}$ 
& 77.12$\pm$0.77 & 63.41$\pm$1.09 & 74.38$\pm$0.45  &  53.79$\pm$1.34 \\
$\mathcal R^{32}_{0}$ & $\mathcal R^{32}_{0}$ & $\mathcal R^{32}_{0}$ 
& 77.80$\pm$0.56 & 64.60$\pm$1.46 & 74.21$\pm$0.86  &  52.93$\pm$1.26 \\
\hline
\end{tabular}}
\end{table}

\begin{figure}[htbp]
\vspace{-5pt}
  \centering
  \begin{subfigure}[t]{0.49\linewidth}
    \centering
    \includegraphics[width=\linewidth]{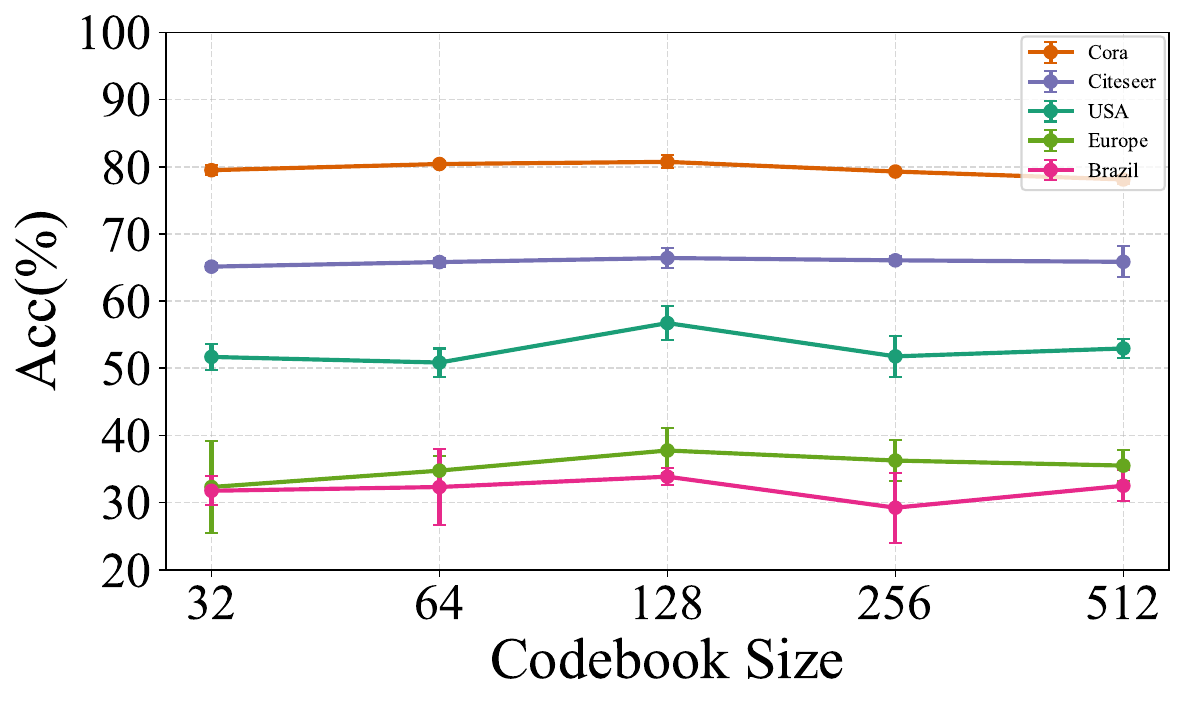}
    \caption{Node classification results.}
    \label{fig2a}
    \vspace{-10pt}
  \end{subfigure}
  \hfill
  \begin{subfigure}[t]{0.49\linewidth}
    \centering
\includegraphics[width=\linewidth]{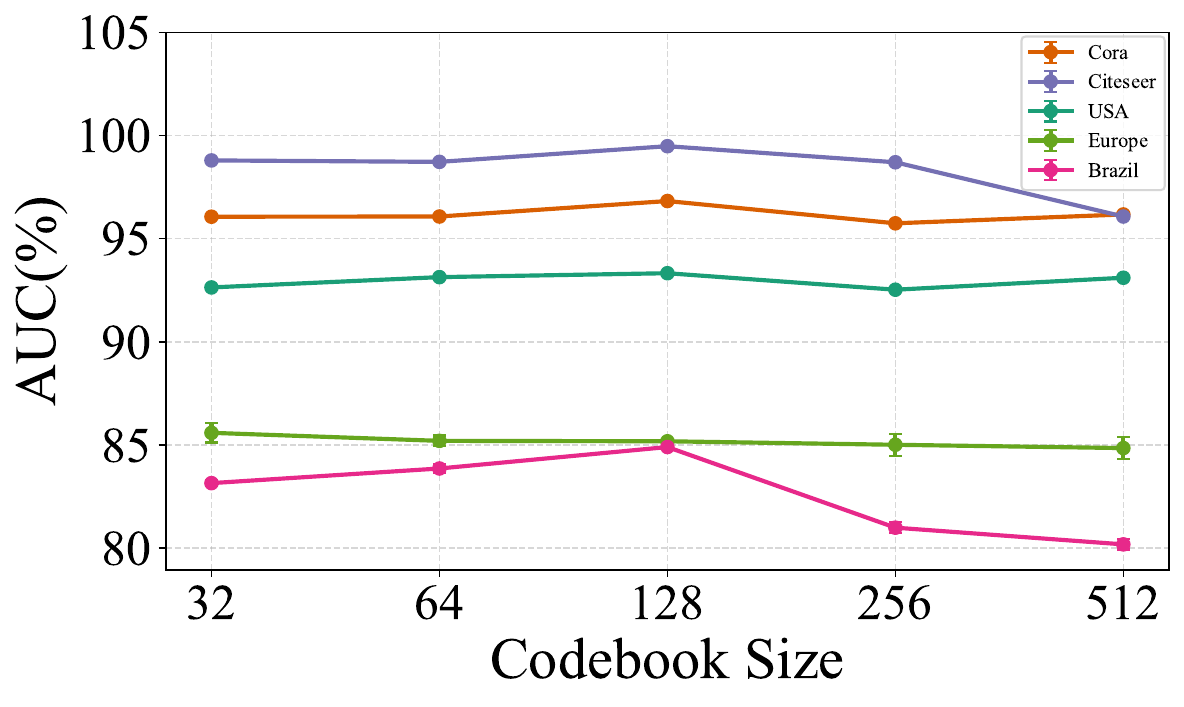}
    \caption{Link prediction results.}
    \vspace{-10pt}
    \label{fig2b}
  \end{subfigure}
  \caption{Hyperparameter sensitivity analysis of codebook size.}
  \label{fig2}
  \vspace{-10pt}
\end{figure}

\paragraph{Impact of Hyperparameters.}
To analyze the sensitivity of model parameters, we focus on the impact of the codebook size on model performance. For each setting of the $m$, the model is pretrained from scratch and evaluated across multiple datasets. The test results for node classification and link prediction under various datasets are illustrated in Figure~\ref{fig2a} and Figure~\ref{fig2b}. The results indicate that a limited number of structural tokens in the codebook is sufficient to achieve competitive performance. However, when the codebook size becomes too large, performance degrades due to the increased risk of codebook collapse. This highlights the importance of selecting an appropriate codebook size for stable and effective structure-level quantization.

\begin{figure}[htbp]
\vspace{-5pt}
  \centering
  \begin{subfigure}[t]{0.49\linewidth}
    \centering
    \includegraphics[width=\linewidth]{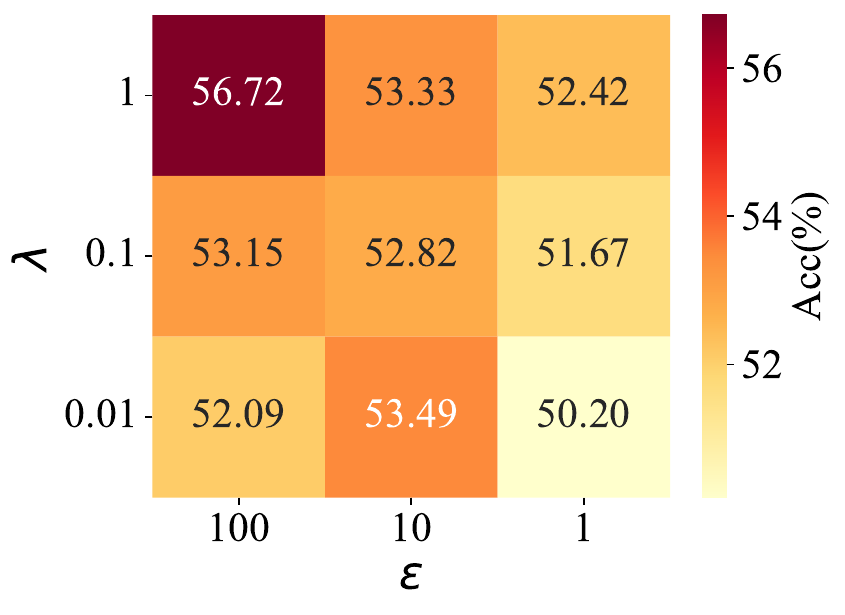}
    \caption{Heatmap about $\lambda$ and $\epsilon$.}
    \vspace{-6pt}
    \label{fig3a}
  \end{subfigure}
  \hfill
  \begin{subfigure}[t]{0.44\linewidth}
    \centering
    \includegraphics[width=\linewidth]{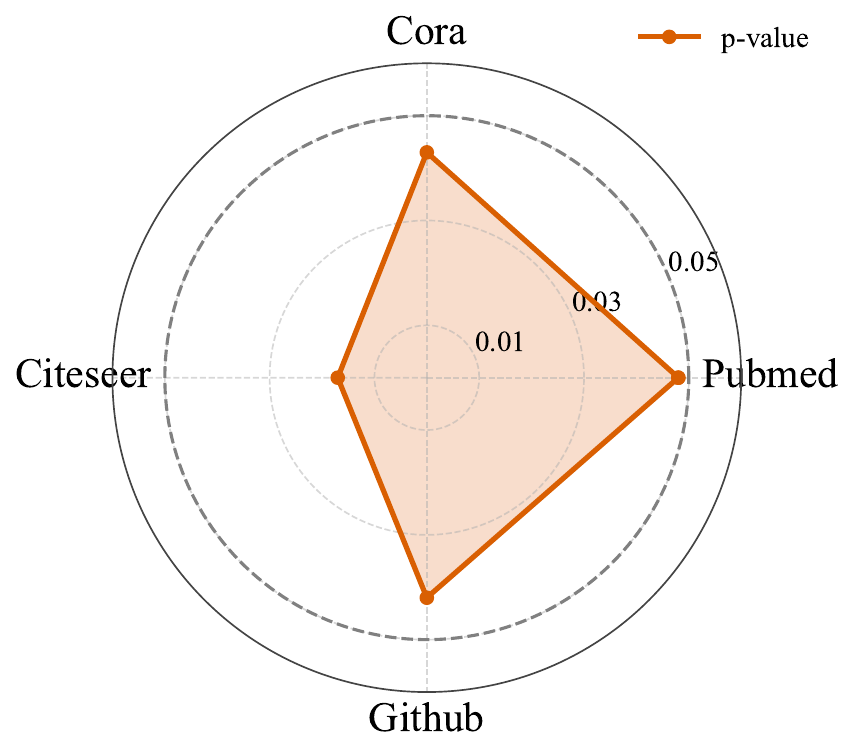}
    \caption{P-value on four datasets.}
    \vspace{-6pt}
    \label{fig3b}
  \end{subfigure}
    \hfill
  \caption{Heatmap about $\lambda$ and $\epsilon$ and radar plot of p-value.}
  \label{fig3}
  \vspace{-10pt}
\end{figure}

We also perform an hyperparameter study about $\lambda, \epsilon$. The results of node classification on USA dataset are shown in the Figure~\ref{fig3a}. Hence, the best $\lambda= 1, \epsilon =100$ in our experiment. 

 \vspace{-5pt}

\paragraph{Statistical Analysis.}
To mitigate the influence of randomness and ensure the statistical reliability of the results, we first systematically evaluate the performance of RGT compared to RiemannGFM across multiple datasets and random seeds.
Five independent experiments are conducted on the Pubmed dataset, and Table~\ref{tab6} reports the mean and standard deviation of Acc and F1.
Paired t-test results show that RGT significantly outperforms RiemannGFM at the 99\% confidence level.
Furthermore, Figure~\ref{fig3b} presents the radar plot of p-value under the five-shot setting across four datasets. The results $p < 0.05 $ demonstrate that RGT significantly outperforms RiemannGFM on certain datasets at the 95\% confidence level.

\vspace{-7pt}
\begin{table}[htbp]
\centering
\caption{Cross-domain transfer learning performance of node classification on Pubmed (Bold: best).}
\vspace{-10pt}
\label{tab6}
\renewcommand{\arraystretch}{1.1}
\resizebox{\columnwidth}{!}{
\begin{tabular}{c|cc|cc}
\hline
\textbf{Seed} & \textbf{RiemannGFM (Acc)} & \textbf{RGT (Acc)} & \textbf{RiemannGFM (F1)} & \textbf{RGT (F1)} \\
\hline
Seed 0 & 73.94$\pm$5.66 & \textbf{76.22$\pm$0.80} & 74.25$\pm$4.25 & \textbf{76.14$\pm$0.77} \\
Seed 1 & 71.52$\pm$5.44 & \textbf{75.94$\pm$0.88} & 72.11$\pm$4.01 & \textbf{75.89$\pm$0.73} \\
Seed 2 & 72.03$\pm$5.70 & \textbf{76.08$\pm$0.79} & 73.09$\pm$4.37 & \textbf{76.07$\pm$0.80} \\
Seed 3 & 74.11$\pm$5.61 & \textbf{76.35$\pm$0.85} & 73.64$\pm$4.40 & \textbf{76.25$\pm$0.82} \\
Seed 4 & 69.77$\pm$5.58 & \textbf{75.71$\pm$0.81} & 72.87$\pm$4.16 & \textbf{75.83$\pm$0.75} \\
\hline
\textbf{Mean} & 72.67$\pm$5.60 & \textbf{76.06$\pm$0.83} & 73.19$\pm$4.24 & \textbf{76.04$\pm$0.77} \\
\textbf{p-value} & -- & \textbf{0.0056} & -- & \textbf{0.0008} \\
\hline
\end{tabular}}
\vspace{-10pt}
\end{table}
\vspace{-8pt}

\begin{figure}[htbp]
  \centering
  \begin{subfigure}[t]{0.49\linewidth}
    \centering
    \includegraphics[width=\linewidth]{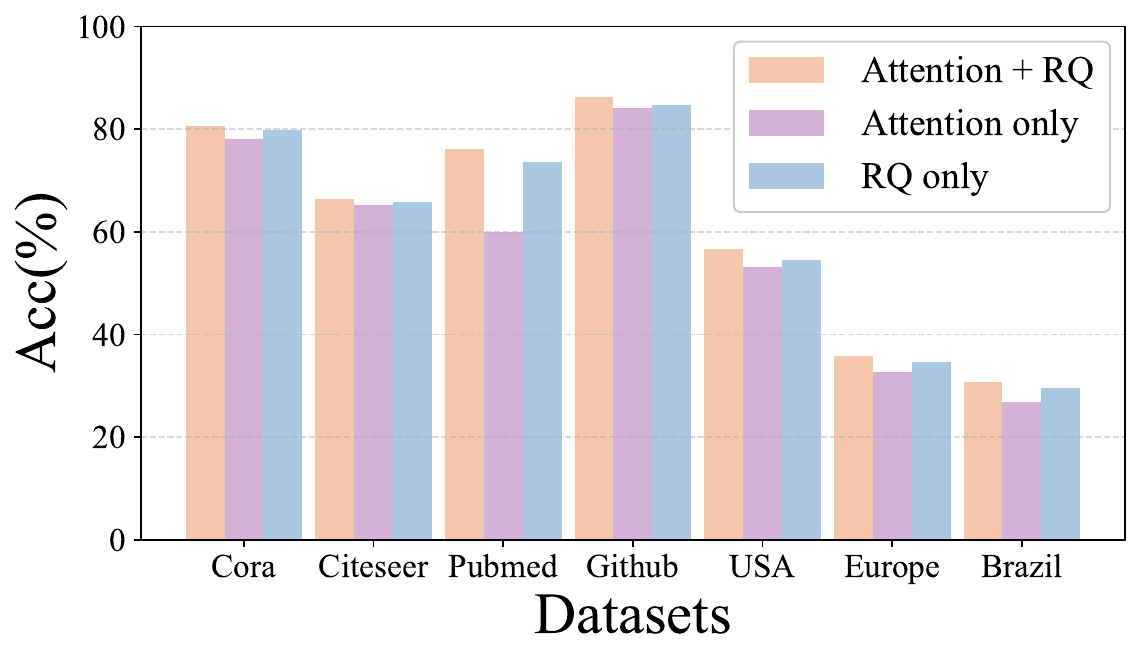}
    \caption{Acc of ablation models}
    \label{fig4a}
    \vspace{-6pt}
  \end{subfigure}
  \hfill
  \begin{subfigure}[t]{0.49\linewidth}
    \centering
    \includegraphics[width=\linewidth]{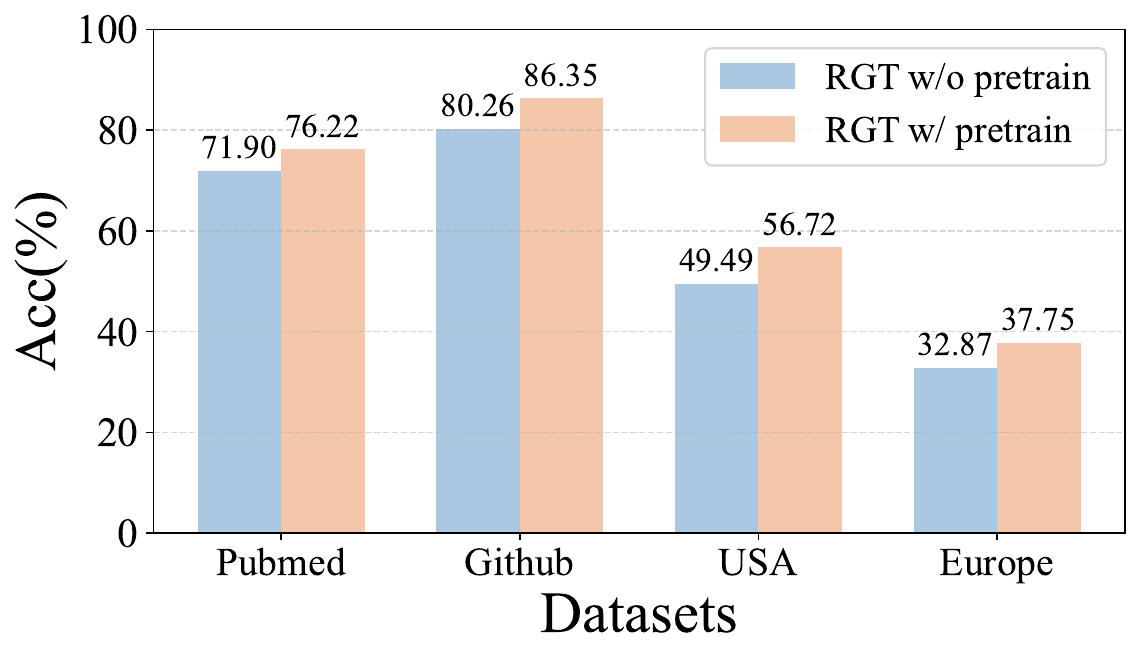}
    \caption{Ablation on RGT Pretraining.}
    \label{fig4b}
    \vspace{-6pt}
  \end{subfigure}
  \caption{Ablation study on key components.}
  \label{fig4}
  \vspace{-10pt}
\end{figure}

\paragraph{Ablation Study.}
The core components of RGT are the cross-geometry attention and the Riemannian quantizer, which are responsible for encoding local coordinates and globally quantizing structural features, respectively. Therefore, it is essential to conduct ablation studies on these two components. The link prediction results across various datasets are presented in Figure~\ref{fig4a}. The results show that applying Riemannian quantizer leads to better performance compared to the non-quantized variant. In addition, the cross-geometry attention mechanism also contributes positively to the results. Overall, Riemannian quantizer plays a more critical role in enhancing the model’s transferability across domains.

To further demonstrate the effectiveness of cross-domain structural knowledge transfer, we compare the performance of RGT with and without pretraining.
The node classification results are presented in Figure~\ref{fig4b}, where the variant without pretraining is labeled as “w/o pretrain.”
The results clearly show that pretraining significantly enhances RGT’s capability in transferring structural knowledge across domains.

 \vspace{-5pt}

\paragraph{Impact of Pretraining Datasets(RQ4).}
This part investigates the impact of different training datasets on RGT. The are pretrained on three graph datasets (Flickr, AComp and WikiCS) respectively and the results of link prediction are reported in Table~\ref{tab7}. The results indicate that the domain and number of datasets used for pretraining can influence model performance. RGT and RiemannGFM maintain stable performance under cross-domain settings. In contrast, while OpenGraph performs well when trained and tested on WikiCS (a citation network), its performance on the Citeseer drops significantly when trained on Flickr or AComp. Moreover, OpenGraph is unable to handle non-attributed graphs, which further limits its transfer ability across domains. 
\vspace{-10pt}

\begin{table}[htbp]
        \caption{Cross-domain link prediction performance on different pretraining datasets.}
        \vspace{-10pt}
        \label{tab7}
        \centering
        \LARGE
\renewcommand{\arraystretch}{1}
\resizebox{\columnwidth}{!}{
        \begin{tabular}{c  c | c c c c}
        \hline
        \textbf{Pretraining} & \textbf{Method}  & \textbf{Citeseer} & \textbf{Cora}  & \textbf{USA}  & \textbf{Brazil} \\
        \hline
        \multirow{3}{*}{Flickr}
        & OpenGraph 
        & 73.17{$\pm$0.23} 
        & 77.46{$\pm$0.23} 
        & 53.16{$\pm$0.98}
        & 61.16{$\pm$0.19}\\
        & RiemannGFM 
        & 98.27{$\pm$1.38}
        & 96.84{$\pm$0.24} 
        & 93.40{$\pm$0.84}
        & 84.03{$\pm$0.26}\\
        & \textbf{RGT}(Ours)  &  \textbf{99.11{$\pm$0.01}} &\textbf{97.81{$\pm$0.32} }& \textbf{93.41{$\pm$0.02}} 
        & \textbf{84.92{$\pm$0.50}}\\
        \hline
        \multirow{3}{*}{AComp}
        & OpenGraph 
        & 72.95{$\pm$0.28} 
        & 78.47{$\pm$0.25} 
        & 79.79{$\pm$0.81}
        & 68.31{$\pm$0.14}\\
        & RiemannGFM &   98.50{$\pm$0.01}&96.59{$\pm$0.00} & 92.85{$\pm$0.13} 
        & 84.32{$\pm$0.13}\\
          & \textbf{RGT}(Ours) &  \textbf{98.87{$\pm$0.03}} &\textbf{97.91{$\pm$0.17}} & \textbf{92.90{$\pm$0.02} }
          & \textbf{84.89{$\pm$0.11}}\\
        \hline
        \multirow{3}{*}{WikiCS}
        & OpenGraph 
        & 87.01{$\pm$1.26}
         & 78.07{$\pm$0.12}
        & 80.72{$\pm$1.45} 
        & 63.57{$\pm$0.74} \\
        & RiemannGFM &   98.44{$\pm$0.81}&96.62{$\pm$0.02} 
         & 92.63{$\pm$0.05}
          & 85.10{$\pm$0.08}\\
        & \textbf{RGT}(Ours)  &  \textbf{98.81{$\pm$0.02}}&\textbf{97.56{$\pm$0.08}} & \textbf{92.63{$\pm$0.05}}
         & \textbf{85.91{$\pm$0.54}}\\
        \hline
        \end{tabular}}
        \end{table}    
\vspace{-10pt}

\paragraph{Scale to Web-scale Graph(RQ5).}

\begin{figure}[htbp]
  \centering
\includegraphics[width=0.49\linewidth]{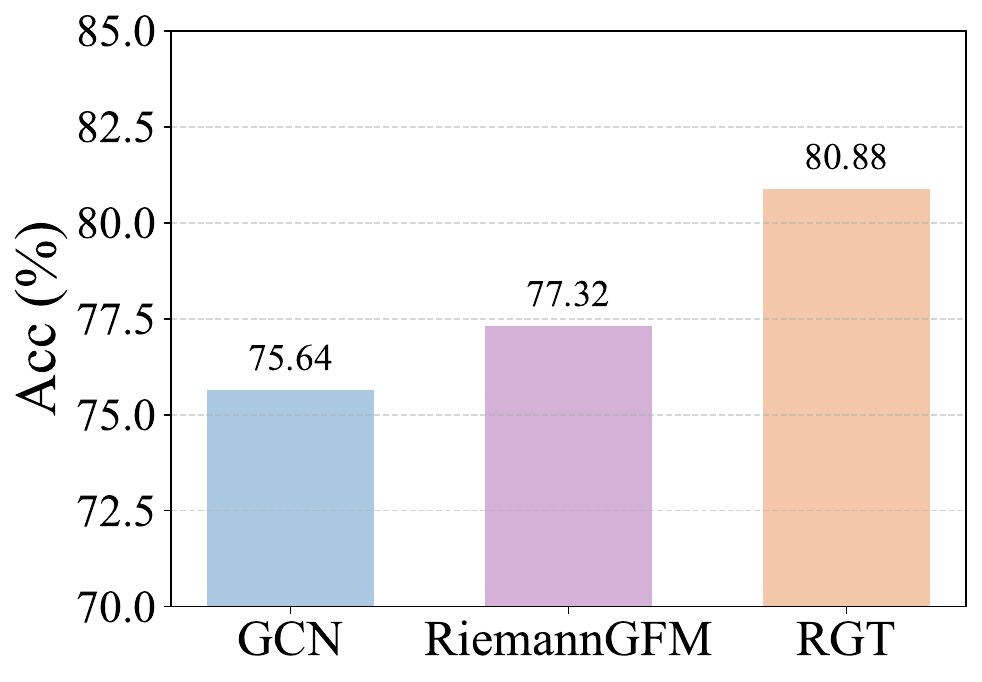}
  \caption{Cross-domain Node Classification Performance on Web-Scale Graphs.}
  \vspace{-10pt}
  \label{fig5}
\end{figure}

We evaluate the performance of the pretrained RGT model on the ogbn-products dataset. The cross-domain node classification results are illustrated in Figure~\ref{fig5}. During inference, the primary computational overhead originates from eigenvalue computation and subgraph sampling.
Nevertheless, RGT remains capable of efficiently handling datasets of this scale. These results demonstrate that RGT exhibits strong scalability and competitive performance on web-scale graph datasets.

\vspace{-10pt}
 \vspace{-5pt}

\section{Conclusion}\label{sec5}
This work primarily contributes a novel Riemannian graph tokenizer, designed to quantise and transfer structural knowledge across graph domains. Specifically, we first encode the coordinates of geometric vocabulary. Subsequently, the proposed Riemannian quantizer facilitates the learning of structural knowledge tranfer. During the quantization stage, we propose a Riemannian Straight-Through Estimator (RSTE) to maintain the intrinsic structure. The proposed model is evaluated on both attributed and non-attributed graph datasets. The empirical results demonstrate the superiority of RGT. Future work may explore jointly encoding structural semantics and textual features.



\section{Acknowledgments}
This work is supported in part by the National Natural Science Foun-dation of China(No.62550138,62192784,62572064,62472329,62202164).




\bibliographystyle{ACM-Reference-Format}
\bibliography{RGT}
\appendix

\vspace{-5pt}
\section{Theoretical Analysis}\label{C}

\subsection{The Completeness of Geometric Vocabulary}\label{C.1}
Definition~\ref{def1} formalizes a complete geometric vocabulary that supports the structural patterns of arbitrary graph.


From the perspective of graph theory, any graph $\mathcal{G} = (\mathcal{V}, \mathcal{E})$ can be fully described as a combination of tree and cycle structures. A spanning tree (composed by stitching together multiple subtrees) provides the connected backbone of the graph, while the non-tree edges form cycles. Therefore, the combination of \textbf{trees} and \textbf{cycles} is sufficient to reconstruct the graph topology in terms of connectivity.

However, from the geometric modeling and topological structure perspective, it is necessary to additionally introduce the \textbf{sequence structure} as a linear substructure. A sequence represents a segment of the graph with zero curvature, i.e., a locally flat subgraph that serves as a first-order Euclidean approximation. The geometric vocabulary possesses the following geometric interpretations:
\newenvironment{paraitem}{\begin{itemize}}{\end{itemize}}
\begin{itemize}
    \item Tree structures are mapped to hyperbolic space ($\kappa < 0$), capturing hierarchical and branching patterns,
    \item Cycle structures are mapped to spherical space ($\kappa > 0$), encoding closed-loop topologies,
    \item Sequence structures are mapped to Euclidean space ($\kappa = 0$), modeling locally linear or path-like substructures.
\end{itemize}

\begin{figure}[htbp]
\vspace{-8pt}
  \centering
  \includegraphics[width=0.6\linewidth]{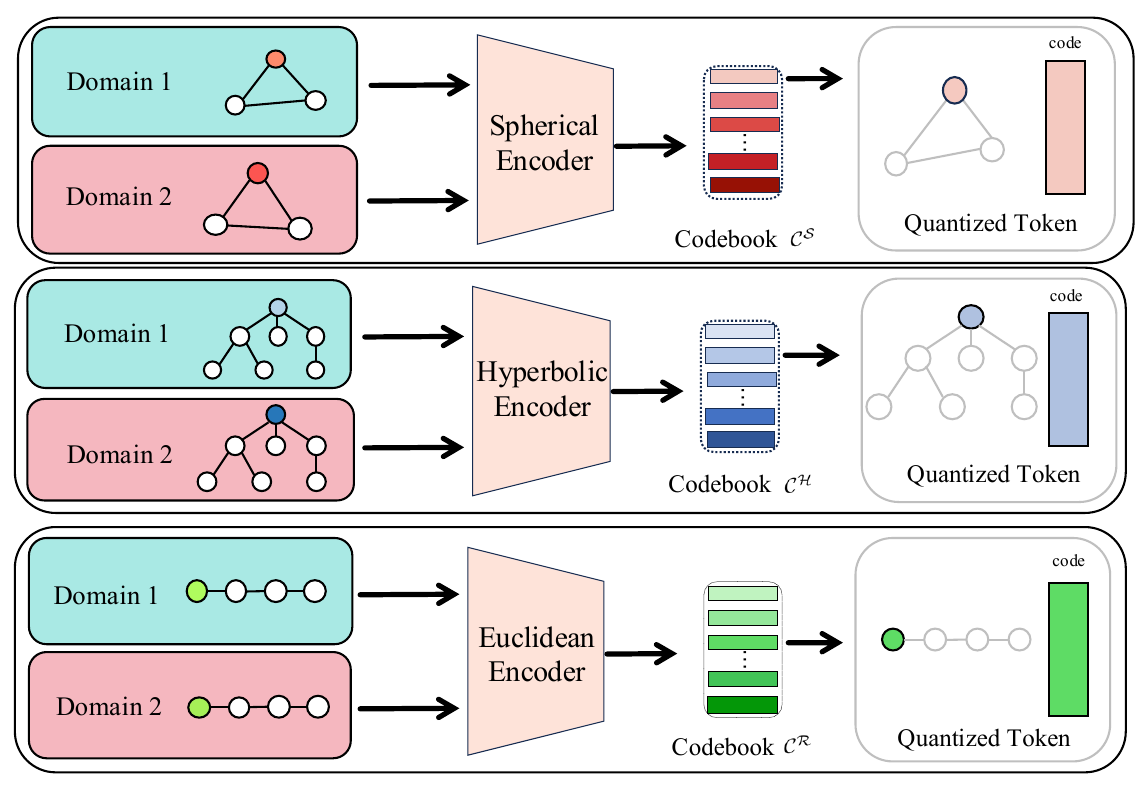}
  \caption{Cross-Domain Transfer of Geometric Vocabulary.}
  \label{fig:geovoc}
\vspace{-10pt}
\end{figure}

Therefore, the trees, cycles and sequences constitute a complete geometric vocabulary for representing graph structures. Each structural type is naturally associated with a constant curvature space—hyperbolic (negative), spherical (positive), or Euclidean (zero). They provide a unified framework for modeling arbitrary graphs across diverse structural patterns, topologies and geometric characteristics. Figure~\ref{fig:geovoc} illustrates the cross-domain transferability of RGT based on geometric vocabulary.


\subsection{Proof of The Theorem 1}\label{C.2}

Let $c_i^{\mathcal{M}} = RQ(z_i^{\mathcal{M}}), c_j^{\mathcal{M}} = RQ(z_j^{\mathcal{M}})$. By the triangle inequality, we have $d(c_i^{\mathcal{M}}, c_j^{\mathcal{M}}) \le d(c_i^{\mathcal{M}}, z_i^{\mathcal{M}}) + d(z_i^{\mathcal{M}}, z_j^{\mathcal{M}})+ d(z_j^{\mathcal{M}}, c_j^{\mathcal{M}})$. Since $ c_i^{\mathcal{M}}, c_j^{\mathcal{M}} $ are the nearest codewords of $ z_i^{\mathcal{M}}, z_j^{\mathcal{M}} $ and $ d(z_i^{\mathcal{M}}, z_j^{\mathcal{M}}) \le \epsilon $. Then we obtain
$d(c_i^{\mathcal{M}}, z_i^{\mathcal{M}}) \le \epsilon, \quad d(c_j^{\mathcal{M}}, z_j^{\mathcal{M}}) \le \epsilon$. Hence,  $d(c_i^{\mathcal{M}}, c_j^{\mathcal{M}}) \le 3\epsilon$, the theorem is proved.

\subsection{Proof of The Theorem 2}\label{C.3}

Let $(\mathcal{M}, g)$ be a smooth Riemannian manifold equipped with a geodesic distance function $d_{\mathcal{M}}(\cdot,\cdot)$. Consider a set of geometric vocabulary $\{\mathcal{S}_i^{(j)}\}_{j=1}^M$ and a codebook $\mathcal{C} = \{\boldsymbol{c}_j^{\mathcal{M}}\}_{j=1}^m \subset \mathcal{M}^d$. The quantizer $RQ: \{\mathcal{S}_i^{(j)}\}_{j=1}^M \to \mathcal{C}$ assigns each geometry vocabulary $\boldsymbol{z}$ to a codeword according to:

\begin{equation}\label{eq24}
\boldsymbol{c}_j^{\mathcal{M}} = \arg\min_{\boldsymbol{c} \in \mathcal{M}^d} \mathbb{E}\left[ d_{\mathcal{M}}(\boldsymbol{z}, \boldsymbol{c}) \mid \boldsymbol{z} \in \{\mathcal{S}_i^{(j)}\}_{j=1}^M \right],
\end{equation}

Then, the quantizer $RQ$ minimizes the expected geodesic distortion over the entire space:
\begin{equation}\label{eq25}
\mathbb{E}\left[ d_{\mathcal{M}}(\boldsymbol{z}, RQ(\boldsymbol{z})) \right]. 
\end{equation}

Let $P_j = \mathbb{P}(\boldsymbol{z} \in \{\mathcal{S}_i^{(j)}\}_{j=1}^M)$ denote the probability that a geometry vocabulary falls into region $\{\mathcal{S}_i^{(j)}\}_{j=1}^M$. Suppose $RQ'$ is another quantizer defined on the same set $\{\mathcal{S}_i^{(j)}\}_{j=1}^M$, but with a different codebook $\mathcal{C}' = \{\boldsymbol{c}_j'\}_{j=1}^m$. The expected distortion of $RQ'$ is given by:
\begin{align}\label{eq26}
\mathbb{E}[d_{\mathcal{M}}(\boldsymbol{z}, RQ'(\boldsymbol{z}))] 
&= \sum_{j=1}^m P_j \cdot \mathbb{E}[d_{\mathcal{M}}(\boldsymbol{z}, RQ'(\boldsymbol{z})) \mid \boldsymbol{z} \in \{\mathcal{S}_i^{(j)}\}_{j=1}^M] \notag \\
&= \sum_{j=1}^m P_j \cdot \mathbb{E}[d_{\mathcal{M}}(\boldsymbol{z}, \boldsymbol{c}_j') \mid \boldsymbol{z} \in \{\mathcal{S}_i^{(j)}\}_{j=1}^M] \notag \\
&\geq \sum_{j=1}^m P_j \cdot \min_{\boldsymbol{c} \in \mathcal{C}} \mathbb{E}[d_{\mathcal{M}}(\boldsymbol{z}, \boldsymbol{c}) \mid \boldsymbol{z} \in \{\mathcal{S}_i^{(j)}\}_{j=1}^M] \notag \\
&= \sum_{j=1}^m P_j \cdot \mathbb{E}[d_{\mathcal{M}}(\boldsymbol{z}, \boldsymbol{c}_j^{\mathcal{M}}) \mid \boldsymbol{z} \in \{\mathcal{S}_i^{(j)}\}_{j=1}^M] \notag \\
&= \mathbb{E}[d_{\mathcal{M}}(\boldsymbol{z}, RQ(\boldsymbol{z}))]. 
\end{align}
\paragraph{Conclusion.} Therefore, choosing each codeword $\boldsymbol{c}_j^{\mathcal{M}}$ as the global centroid of $\{\mathcal{S}_i^{(j)}\}_{j=1}^M$ across multiple domains minimizes the total expected geodesic distortion. This confirms that the quantizer $RQ$ defined in Eq.(\ref{eq6}) is the optimal structure-level Riemannian quantization strategy.

\section{Riemannian Straight-Through Estimation and Codebook Collapse}\label{D}

\subsection{Riemannian Straight-Through Estimator}\label{D.1}

Let $\boldsymbol z_i^{\mathcal{M}}$ denote the encoder output in manifold $\mathcal{M}$ and $\boldsymbol q_i^{\mathcal{M}}$ denote the corresponding nearest code. Since the nearest-neighbor quantization operation is non-differentiable, we adopt the following Riemannian Straight-Through Estimation (RSTE) formulation:

\begin{equation}\label{eq27}
\hat{\boldsymbol z_i}^{\mathcal{M}} = \exp_{\boldsymbol z_i^{\mathcal{M}}}\left( \mathrm{sg}\left[ \log_{\boldsymbol z_i^{\mathcal{M}}}(\boldsymbol q_i^{\mathcal{M}}) \right] \right), \quad \boldsymbol q_i^{\mathcal{M}} \in \mathcal{M},
\end{equation}
where $\log_{\boldsymbol z_i^{\mathcal{M}}}(\boldsymbol q_i^{\mathcal{M}})$ maps the code $\boldsymbol q_i^{\mathcal{M}}$ onto the tangent space at $\boldsymbol z_i^{\mathcal{M}}$. $\mathrm{sg}[\cdot]$ denotes the stop-gradient operator, which ensures that gradients are only propagated through $\boldsymbol z_i^{\mathcal{M}}$ and not through $\boldsymbol q_i^{\mathcal{M}}$.
\vspace{-5pt}
\subsection{RSTE Alleviates Codebook Collapse}\label{D.2}

Codebook collapse is a major challenge in vector quantization (VQ) methods, and numerous studies have sought to address this issue (e.g., SimVQ, RVQ and other related works). Codebook collapse commonly occurs in the following scenarios. When the encoded vector $\boldsymbol z$ is significantly distant from its assigned discrete code $\boldsymbol q$, applying the Euclidean interpolation $\boldsymbol z + \mathrm{sg}[\boldsymbol q - \boldsymbol z]$ tends to pull multiple $z$ vectors into a shared local Euclidean neighborhood. During training, gradients are propagated purely in Euclidean space, causing the optimization to overfit to a small subset of codebook entries, while leaving other codes unused or inactive. In contrast, RVQ leverages the Riemannian structure by computing the local geometric offset of the code $q$ relative to $\boldsymbol z$ via the logarithmic map $\log_{\boldsymbol z}(\boldsymbol q)$. During the exponential map $\exp_{\boldsymbol z}(\cdot)$ that projects back to the manifold, each codeword maintains its manifold-aware structure. Gradients are propagated within the tangent space $\mathcal{T}_{\boldsymbol z} \mathcal{M}$, which is consistent with the underlying geometry of the space. Alternatively, other techniques such as expiring stale codes ~\cite{zeghidour2021soundstream} or affine re-parameterization ~\cite{huh2023straightening} can also be employed to mitigate this problem. As a result, different structural samples are encouraged to select codes centered around different geometric regions of the manifold. This promotes diversified codeword utilization and effectively mitigates the codebook collapse issue.

\section{Algorithm}\label{E}
The overall training process of RGT is presented in Algorithm 1.
\vspace{-5pt}
\begin{algorithm}[htbp]
        \caption{Training Algorithm of RGT} 
        \label{Al1}
        \KwIn{multi-domain graphs, Hyperparameters of RGT.}
        \KwOut{Model parameters of RGT}
        Initialize node features and node coordinates on CCSs;
        Sample substructures according to the geometry vocabulary;

\While{model not converged}{   
            \For{each substructure in each geometry} {
               Conduct the cross-geometry attention in Eq.(\ref{eq2}) and update node coordinates by Eq.(\ref{eq3})

                Conduct the Riemannian Quantizer in Eq.(\ref{eq6}) to tokenise graph structures
                
                Conduct the geometry-aligned decoder in Eq.(\ref{eq10})}
            Compute the loss with Eq.(\ref{eq11})

            Update model parameters via gradient descent via Eq.(\ref{eq15}) and Eq.(\ref{eq16}).
}
\end{algorithm}

\section{Experiment Details}\label{F}

\subsection{Experimental Setup}\label{F.3}
During the experimental process, we implement each layer of the model as described in the section~\ref{sec3}. Two aspects of the setup require special attention. \textbf{Hyperparameter settings:} The number of layers for tree-based sampling is set to 2, cycles are constructed as triangles (3-node), and the sequence length is set to $L$. The embedding dimension $d$ for structural embedding is 32. The curvatures of the constant curvature spaces $\kappa$ are set to 1 (spherical), $-1$ (hyperbolic), and 0 (Euclidean space). The feature dimension of each code is 32, the number of code in codebook $m=128$. All hyperparameters are selected via grid search. For baselines, the hyperparameters strictly follow those reported in the original papers. \textbf{Pretraining and Evaluation:} The pretraining datasets used for our model is the same as in RiemannGFM. Other baselines use their original pretraining datasets. During evaluation, the pretrained codebooks are kept frozen while the GNN classifier is fine-tuned on the target test datasets. Other models are evaluated on the same test datesets. The results are averaged over multiple runs to reduce the impact of randomness. 

\paragraph{Pretraining of RGT}
We pretrain RGT using the Adam optimizer with a learning rate of $0.001$, a batch size of 32. The pretraining stage runs for 3 epochs and 3 iterations. For RQ, the codebook size to 128, the code dimension to 32. Dropout is set to 0.1.

\paragraph{Fine-tuning of RGT}
We fine-tune RGT for node classification and link prediction. For node classification, we use a learning rate of $0.01$ and train for 120 epochs. The task hidden dimension is set to 128. For link prediction, we train for 120 epochs with a hidden dimension of 256. Drop edge and drop feature rates are set to 0.2 and 0.3, respectively.

\begin{table}[htbp]
\centering
\caption{Hyperparameters for RGT in pretraining and fine-tuning}
\label{tab:rgt-hyper}

\renewcommand{\arraystretch}{0.9} 
\setlength{\tabcolsep}{4pt}       

\begin{tabular}{l|c|c}
\hline
\textbf{Hyperparameter} & \textbf{Pretraining} & \textbf{Fine-tuning} \\
\hline
Learning Rate         & 0.001 & 0.01 \\
Epochs                & 3     & 120  \\
Batch Size            & 32    & 64   \\
Dropout               & 0.1   & 0.1  \\
Drop Edge Rate        & --    & 0.2  \\
Drop Feature Rate     & --    & 0.3  \\
Embedding Dimension   & 32    & 32   \\
Hidden Dimension      & 256   & 256  \\
Task Hidden Dimension & --    & 128  \\
Codebook Size         & 128   & --   \\
Code Dimension        & 32    & --   \\
Patience              & 3     & 10   \\
\hline
\end{tabular}
\end{table}
\vspace{-10pt}


\paragraph{Comparison with GFT and VQGraph.}
RGT, GFT~\cite{DBLP:conf/nips/WangZCZ024} and VQGraph~\cite{DBLP:conf/iclr/0006TXLHQZ0ZL24} all leverage vector quantization techniques to extract discrete structural semantics from graphs. However, they differ significantly in terms of core design principles and application scenarios. The key distinctions between RGT and the other two models are summarized as follows:

\vspace{-10pt}
\begin{table}[htbp]
    \caption{Cross-domain transfer learning performance of node classification on three datasets.}
    \vspace{-10pt}
    \label{tab10}
    \centering
    \renewcommand{\arraystretch}{0.85}
    \resizebox{\columnwidth}{!}{
    \begin{tabular}{c| cc  cc  cc }
    \hline
    & \multicolumn{6}{c}{\textbf{Node Classification Results}}  \\ 
    \cline{2-7}
    \textbf{Method} 
    & \multicolumn{2}{c}{\textbf{Cora}}  
    & \multicolumn{2}{c}{\textbf{Citeseer}} 
    & \multicolumn{2}{c}{\textbf{Brazil}} \\
    \cline{2-7}
    & ACC & F1 
    & ACC & F1 
    & ACC & F1\\
    \hline
    DGI
    & 80.62{$\pm$1.63} & 80.67{$\pm$1.97} 
    & 69.67{$\pm$0.42} & 62.92{$\pm$1.81} 
    & 31.78{$\pm$1.62} & 29.15{$\pm$4.16} \\
    GraphMAE
    & 79.77{$\pm$0.34} & 79.82{$\pm$0.23} 
    & 50.20{$\pm$7.07} & 44.97{$\pm$8.20} 
    & 32.14{$\pm$5.83} & 21.14{$\pm$7.72} \\
    OFA
    & 56.92{$\pm$3.09} & 54.21{$\pm$2.02} 
    & 49.68{$\pm$1.36} & 50.22{$\pm$2.13} 
    & - & - \\
    ZeroG
    & 59.20{$\pm$1.07} & 55.32{$\pm$1.34} 
    & 52.34{$\pm$0.89} & 55.67{$\pm$1.22} 
    & - & - \\
    Graphany
    & 79.98{$\pm$1.57} & 77.56{$\pm$1.13} 
    & 63.73{$\pm$1.35} & 62.94{$\pm$1.32} 
    & 29.24{$\pm$1.24} & 20.24{$\pm$2.48} \\
    OpenGraph 
    & 75.03{$\pm$0.97} & 74.37{$\pm$0.80} 
    & 62.34{$\pm$1.06} & 60.29{$\pm$1.34} 
    & 20.42{$\pm$2.33} & 15.63{$\pm$1.31} \\
    GFT
    & 73.38{$\pm$1.03} & 73.83{$\pm$1.08} 
    & - & - 
    & - & - \\
    RiemannGFM
    & 78.66{$\pm$0.47} & 79.01{$\pm$0.34} 
    & 64.78{$\pm$2.01} & 63.95{$\pm$1.69} 
    & 30.77{$\pm$4.21} & 22.52{$\pm$4.60} \\
    \textbf{RGT}(Ours)
    & \textbf{80.72{$\pm$0.97}} &\textbf{ 80.79{$\pm$0.92} }
    & \textbf{66.40{$\pm$1.53}} & \textbf{67.81{$\pm$0.94} }
    & \textbf{33.85{$\pm$1.26}} & \textbf{29.27{$\pm$0.00}} \\
    \hline
    \end{tabular}}
    \vspace{-10pt}
\end{table}

\vspace{-10pt}
\begin{table}[htbp]
    \caption{Cross-domain transfer learning performance of link prediction on three datasets.}
    \vspace{-10pt}
    \label{tab11}
    \centering
    \renewcommand{\arraystretch}{0.85}
    \resizebox{\columnwidth}{!}{
    \begin{tabular}{c| cc  cc  cc }
    \hline
    & \multicolumn{6}{c}{\textbf{Link Prediction Results}}  \\ 
    \cline{2-7}
    \textbf{Method} 
    & \multicolumn{2}{c}{\textbf{Cora}}  
    & \multicolumn{2}{c}{\textbf{Citeseer}} 
    & \multicolumn{2}{c}{\textbf{Brazil}} \\
    \cline{2-7}
    & AUC & AP 
    & AUC & AP  
    & AUC & AP \\
    \hline
    DGI
    & 66.06{$\pm$1.66} & 65.64{$\pm$1.91} 
    & 82.81{$\pm$1.12} & 84.72{$\pm$1.15} 
    & 64.09{$\pm$5.91} & 59.09{$\pm$7.17} \\
    GraphMAE
    & 54.86{$\pm$0.20} & 57.87{$\pm$1.58} 
    & 79.18{$\pm$1.02} & 75.51{$\pm$0.94} 
    & 74.00{$\pm$2.81} & 71.86{$\pm$4.04} \\
    OFA
    & 70.84{$\pm$2.01} & 71.34{$\pm$1.89} 
    & 75.61{$\pm$2.34} & 76.13{$\pm$1.26} 
    & - & - \\
    OpenGraph 
    & 74.03{$\pm$0.97} & 73.21{$\pm$0.84} 
    & 74.45{$\pm$0.46} & 76.32{$\pm$0.76} 
    & 65.31{$\pm$1.54} & 64.39{$\pm$1.65} \\
    RiemannGFM
    & 96.75{$\pm$0.14} & 96.07{$\pm$0.15} 
    & 98.85{$\pm$0.21} & 99.04{$\pm$0.08} 
    & 84.18{$\pm$0.20} & 82.38{$\pm$0.11} \\
    \textbf{RGT}(Ours)
    & \textbf{96.82{$\pm$0.01}} & \textbf{96.26{$\pm$0.01} }
    & \textbf{99.48{$\pm$0.00}} & \textbf{99.39{$\pm$0.01}} 
    & \textbf{84.89{$\pm$0.11} }& \textbf{83.61{$\pm$0.10}} \\
    \hline
    \end{tabular}}
    \vspace{-15pt}
\end{table}

Model objective: GFT focuses on building a general-purpose reasoning model for diverse graph tasks, emphasizing adaptability across domains and tasks. VQGraph, on the other hand, is designed as a lightweight structure-aware MLP model for efficient node classification. In contrast, RGT aims to facilitate structural knowledge transfer across domains, with an emphasis on geometric consistency and representation generalization.
    
Pretraining strategy: GFT performs large-scale pretraining across various graph tasks and text-attributed graphs to capture domain-invariant patterns. VQGraph trains on a single dataset to extract local structural information. RGT adopts a cross-domain pretraining strategy to encode structural semantics and geometric mappings over multi-domain graphs.
    
Token representation: GFT directly uses discrete tokens as transferable patterns and inputs them into a downstream classifier. VQGraph treats tokens as auxiliary structural knowledge to enhance MLP training. In contrast, RGT encodes structural tokens in multiple geometric spaces and performs unified fusion in a shared tangent space, enabling consistent geometric alignment and cross-space adaptability.

Downstream applicability: GFT is applicable to a broad range of tasks such as node classification and few-shot learning. VQGraph is primarily designed for conventional node classification with a basic pretraining-finetuning pipeline. RGT supports various graph tasks including node classification and link prediction, and demonstrates strong robustness and transferability, particularly on non-attributed graphs and cross-domain settings.

\vspace{-5pt}
\subsection{Additional Results}\label{F.4}

To further assess the generalization capability of the proposed RGT model, we conduct supplementary cross-domain node classification experiments on three additional datasets. As shown in Table \ref{tab10}, RGT consistently achieves performance that is competitive with or superior to that of existing baselines across all datasets. These results align with the main experimental findings presented in the paper, thereby reinforcing the robustness and adaptability of RGT across diverse domains and graph structures. Collectively, the supplementary experiments provide additional empirical support for the effectiveness of RGT in cross-domain transfer learning tasks.

We further evaluate the generalization capability of RGT through link prediction tasks on the same datasets. As shown in Table \ref{tab11}, RGT maintains consistently strong performance across domains, further reinforcing its robustness and adaptability in diverse graph learning scenarios.

To further examine the generalization of RGT under limited supervision, we conduct additional one-shot node classification experiments on the same set of datasets. As reported in Table \ref{tab12}, while performance varies across methods and datasets, RGT remains competitive in low-label regimes, demonstrating its potential for transfer learning even in extremely data-scarce settings.

\begin{table}[!t]
    \caption{One-shot learning performance of node classification on three datasets.}
    \vspace{-10pt}
    \label{tab12}
    \centering
    \renewcommand{\arraystretch}{0.85}
    \resizebox{\columnwidth}{!}{
    \begin{tabular}{c| cc  cc  cc }
    \hline
    & \multicolumn{6}{c}{\textbf{Node Classification Results}}  \\ 
    \cline{2-7}
    \textbf{Method} 
    & \multicolumn{2}{c}{\textbf{Cora}}  
    & \multicolumn{2}{c}{\textbf{Citeseer}} 
    & \multicolumn{2}{c}{\textbf{Brazil}} \\
    \cline{2-7}
    & ACC & F1 
    & ACC & F1 
    & ACC & F1\\
    \hline
    DGI
    & 39.21{$\pm$1.70} & 34.33{$\pm$2.09} 
    & 34.13{$\pm$3.08} & 31.41{$\pm$1.11} 
    & 29.92{$\pm$3.24} & 21.75{$\pm$2.54} \\
    GraphMAE
    & 51.86{$\pm$8.81} & 50.90{$\pm$8.49} 
    & 41.24{$\pm$8.71} & 36.07{$\pm$9.75} 
    & 26.56{$\pm$3.41} & 17.53{$\pm$3.34} \\
    OFA
    & 57.35{$\pm$3.09} & 55.92{$\pm$2.85} 
    & 36.89{$\pm$3.98} & 38.48{$\pm$5.76} 
    & - & - \\
    ZeroG
    & 63.59{$\pm$1.75} & 63.24{$\pm$1.22} 
    & \textbf{60.34{$\pm$1.21}} & \textbf{58.24{$\pm$1.67} }
    & - & - \\
    Graphany
    & 69.39{$\pm$3.29} & 68.99{$\pm$1.11} 
    & 33.92{$\pm$0.03} & 34.04{$\pm$0.02} 
    & 24.34{$\pm$0.72} & 19.34{$\pm$3.68} \\
    OpenGraph 
    & \textbf{74.63{$\pm$1.30} }& \textbf{74.12{$\pm$1.21} }
    & 58.65{$\pm$1.23} & 56.24{$\pm$1.94} 
    & 20.34{$\pm$2.31} & 21.71{$\pm$1.96} \\
    GFT
    & 42.66{$\pm$6.08} & 35.81{$\pm$6.89} 
    & - & - 
    & - & - \\
    RiemannGFM
    & 57.54{$\pm$4.24} & 56.95{$\pm$5.24} 
    & 38.02{$\pm$9.45} & 32.42{$\pm$9.87} 
    & 25.18{$\pm$2.24} & 21.97{$\pm$5.65} \\
    \textbf{RGT}(Ours)
    & 46.04{$\pm$3.91} & 43.27{$\pm$6.40} 
    & 39.40{$\pm$5.25} & 34.64{$\pm$6.33} 
    & \textbf{30.77{$\pm$4.87}} & \textbf{22.72{$\pm$2.81}} \\
    \hline
    \end{tabular}}
    \vspace{-8pt}
\end{table}

\begin{table}[!t]
    \caption{Five-shot learning performance of node classification on three datasets.}
    \vspace{-10pt}
    \label{tab13}
    \centering
    \renewcommand{\arraystretch}{0.85}
    \resizebox{\columnwidth}{!}{
    \begin{tabular}{c| cc  cc  cc }
    \hline
    & \multicolumn{6}{c}{\textbf{Node Classification Results}}  \\ 
    \cline{2-7}
    \textbf{Method} 
    & \multicolumn{2}{c}{\textbf{Cora}}  
    & \multicolumn{2}{c}{\textbf{Citeseer}} 
    & \multicolumn{2}{c}{\textbf{Brazil}} \\
    \cline{2-7}
    & ACC & F1 
    & ACC & F1 
    & ACC & F1 \\
    \hline
    DGI
    & 64.54{$\pm$4.19} & 65.27{$\pm$4.90} 
    & 48.03{$\pm$3.56} & 44.73{$\pm$3.02} 
    & 30.62{$\pm$3.24} & 29.29{$\pm$2.76} \\
    GraphMAE
    & 72.85{$\pm$1.61} & 71.80{$\pm$1.74} 
    & 47.74{$\pm$4.99} & 40.88{$\pm$6.81} 
    & 28.57{$\pm$6.47} & 20.81{$\pm$5.85} \\
    OFA
    & 65.23{$\pm$3.19} & 62.72{$\pm$2.35} 
    & 43.25{$\pm$3.78} & 40.21{$\pm$4.78} 
    & - & - \\
    ZeroG
    & 63.59{$\pm$1.75} & 63.20{$\pm$1.52} 
    & 55.92{$\pm$5.21} & 52.74{$\pm$4.16} 
    & - & - \\
    Graphany
    & 68.50{$\pm$3.29} & 67.24{$\pm$1.96} 
    & 34.67{$\pm$7.48} & 35.02{$\pm$5.51} 
    & 28.66{$\pm$7.65} & 22.31{$\pm$3.27} \\
    OpenGraph 
    & \textbf{74.42{$\pm$1.38}} & \textbf{73.63{$\pm$1.08} }
    & 50.34{$\pm$3.34} & 51.21{$\pm$2.78} 
    & 28.30{$\pm$3.14} & 24.39{$\pm$3.12} \\
    GFT
    & 43.77{$\pm$7.49} & 39.35{$\pm$6.69} 
    & - & -
    & - & - \\
    RiemannGFM
    & 61.15{$\pm$4.34} & 60.55{$\pm$5.20} 
    & 56.35{$\pm$5.24} & 55.74{$\pm$4.62} 
    & 29.57{$\pm$4.68} & 22.48{$\pm$5.16} \\
    \textbf{RGT}(Ours)
    & 65.54{$\pm$3.38} & 65.19{$\pm$3.87} 
    & \textbf{58.30{$\pm$4.49}} & \textbf{57.44{$\pm$4.95} }
    & \textbf{33.60 {$\pm$2.46}} & \textbf{30.47{$\pm$6.05}} \\
    \hline
    \end{tabular}}
    \vspace{-5pt}
\end{table}

\begin{table}[!t]
\caption{Cross-domain node classification performance on different pre-training datasets.}
\vspace{-10pt}
\label{tab14}
\centering
\renewcommand{\arraystretch}{0.85}
\resizebox{\columnwidth}{!}{
\begin{tabular}{c  c | c c c c}
\hline
& & \multicolumn{4}{c}{\textbf{Testing Datasets}}\\
\textbf{Pre-training} & \textbf{Method}  & \textbf{Citeseer} & \textbf{Cora}  & \textbf{USA}  & \textbf{Brazil} \\
\hline
\multirow{3}{*}{Flickr}
& OpenGraph 
& 58.86{$\pm$0.89} 
& 59.43{$\pm$1.60} 
& 25.97{$\pm$4.98}
& 17.04{$\pm$4.12}\\
& Riemannian 
& 65.74{$\pm$2.07}
& 77.62{$\pm$0.78} 
& 51.34{$\pm$1.61}
& 23.85{$\pm$9.29}\\
& \textbf{RGT}(Ours) 
& \textbf{66.82{$\pm$0.75}} 
& \textbf{80.34{$\pm$0.70}}
& \textbf{54.20{$\pm$1.36} }
& \textbf{28.46{$\pm$4.62}}\\
\hline
\multirow{3}{*}{AComp}
& OpenGraph 
& 46.53{$\pm$1.6} 
& 50.68{$\pm$1.29} 
& 25.88{$\pm$3.84}
& 15.56{$\pm$3.99}\\
& Riemannian 
& 66.4{$\pm$1.76} 
& 80.02{$\pm$0.87} 
& 48.66{$\pm$1.49}
& 30.38{$\pm$3.28}\\
& \textbf{RGT}(Ours)  
& \textbf{67.28{$\pm$0.99}} 
& \textbf{80.32{$\pm$0.67}} 
& \textbf{54.79{$\pm$1.85}}
& \textbf{30.77{$\pm$0.00}}\\
\hline
\multirow{3}{*}{WikiCS}
& OpenGraph 
& 53.01{$\pm$1.02} 
& 68.38{$\pm$0.54}
& 26.89{$\pm$5.22} 
& 15.56{$\pm$4.63} \\
& Riemannian 
& 65.68{$\pm$1.87} 
& 78.42{$\pm$1.05} 
& 51.93{$\pm$1.08}
& \textbf{28.85{$\pm$1.14}}\\
& \textbf{RGT}(Ours)  
& \textbf{66.40{$\pm$0.52}} 
& \textbf{80.10{$\pm$0.81}} 
& \textbf{53.45{$\pm$0.90}}
& 28.46{$\pm$4.62}\\
\hline
\end{tabular}}
\vspace{-5pt}
\end{table}

To complete our analysis under low-resource conditions, we include 5-shot node classification results on three additional datasets in Table \ref{tab13}. RGT shows stable and competitive performance, particularly excelling on Citeseer and Brazil, where it surpasses all baselines. These results further confirm RGT’s capability to generalize under limited supervision, aligning well with our main experimental findings.

To further substantiate the main findings, we conduct additional analysis on the cross-domain node classification performance of RGT under different pre-training datasets. As illustrated in Table \ref{tab14}, RGT consistently achieves superior or competitive results across all target datasets, irrespective of the source domain. These results highlight the robustness of RGT’s pre-training framework and its ability to transfer across diverse domains. While baseline methods may perform adequately under certain configurations, RGT demonstrates more stable and transferable representations overall. This additional evidence further confirms the effectiveness of RGT in cross-domain scenarios, complementing the core experiments presented in the main text.

\clearpage
\raggedbottom









\end{document}